\documentclass[11pt]{article}
\usepackage[preprint]{acl}
\usepackage{times}
\usepackage{latexsym}
\usepackage[T1]{fontenc}
\usepackage[utf8]{inputenc}
\usepackage{microtype}
\usepackage{graphicx}
\usepackage{booktabs}
\usepackage{colortbl}
\usepackage{multirow}
\usepackage{array}
\usepackage{amsmath}
\usepackage{amssymb}
\newcolumntype{P}[1]{>{\raggedright\arraybackslash}p{#1}}

\newcommand{\method}{\textsc{SMRC-SD}}
\newcommand{\fullpath}{\textsc{FullPath-SD}}

\title{When Privileged Guidance Misaligns: State-Matched Routing and Contextualized Self-Distillation for Multi-Turn Agents}
\author{Junzhuo Liu, Weiwei Li, Jun Ling, and Peng Wang\thanks{Corresponding author.} \\
  University of Electronic Science and Technology of China \\
  \texttt{junzhuo.cs@gmail.com},
  \texttt{davelee.uestc@gmail.com} \\
  \texttt{cs.lingjun@gmail.com},
  \texttt{p.wang6@hotmail.com}}

\begin{document}
\maketitle

\begin{abstract}
Multi-turn agents trained with sparse outcome rewards receive limited supervision for intermediate decisions. Privileged on-policy self-distillation provides denser guidance by having a synchronized teacher re-score student responses with training-only references. However, the student may take different actions or complete subgoals in another order, reaching states for which the reference provides no applicable continuation. The resulting state--reference mismatch can lead the reference-conditioned teacher to penalize valid actions. We introduce State-Matched Routing and Contextualized Self-Distillation (SMRC-SD), which uses the student's reached state to decide when and how a reference should guide the student. At each turn, SMRC-SD checks state compatibility with the reference, distills only on matched turns, and contextualizes the reference by augmenting it with a reached-state summary and a compatible continuation. At inference, the policy requires neither references nor state matching. Experiments spanning three model sizes show gains over unconditional full-path self-distillation in embodied interaction (ALFWorld), web navigation (WebShop), and tool use (FTRL). FTRL-trained policies also transfer to BFCL and ToolHop without additional training. Code is available at \url{https://github.com/liujunzhuo/SMRC-SD}.
\end{abstract}

\begin{figure}[!t]
\centering
\includegraphics[width=0.98\columnwidth]{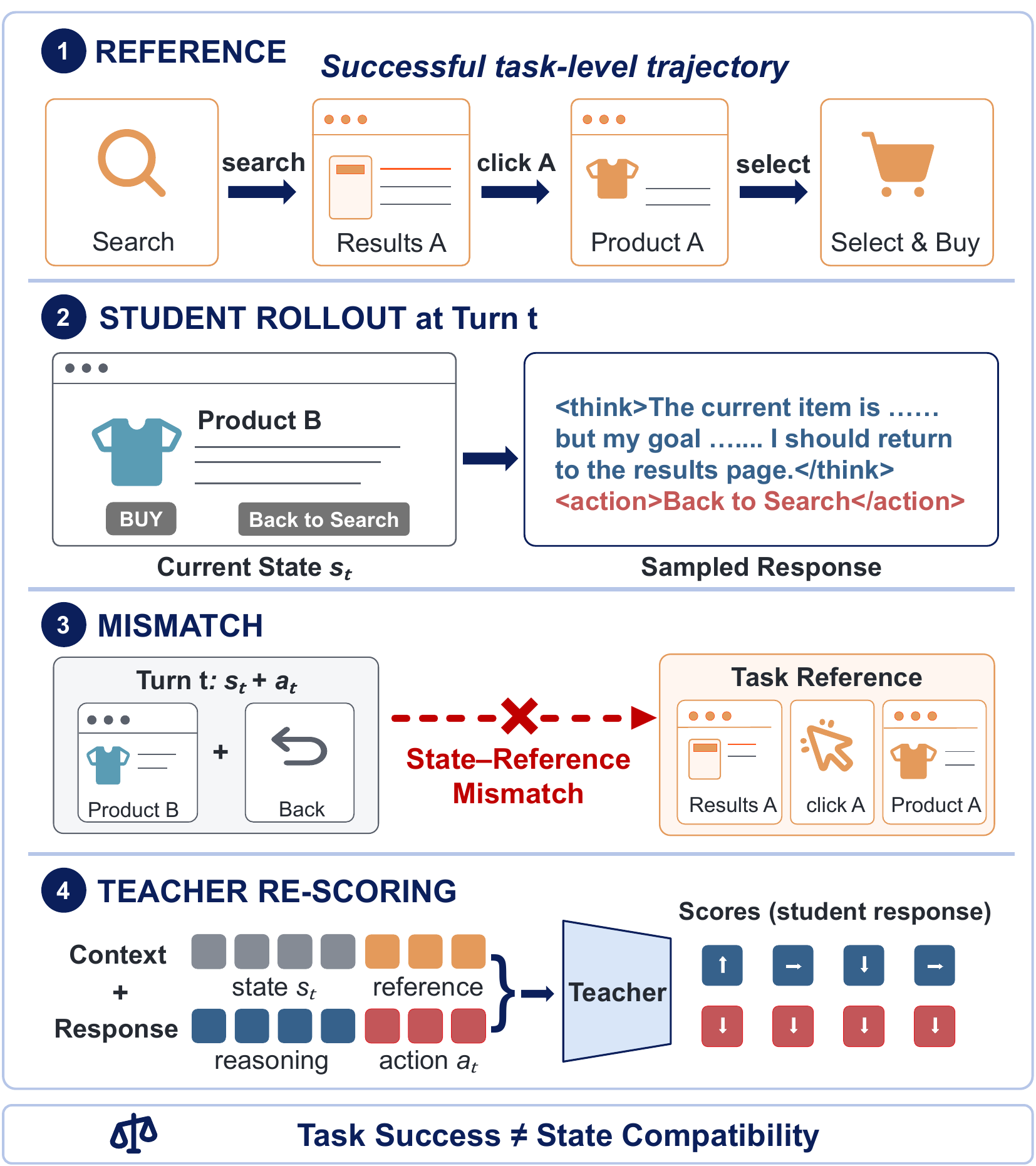}
\caption{State--reference mismatch in Privileged OPD: the student on Product B needs \texttt{Back to Search}, but the reference continues from Results A to Product A. \fullpath{} can suppress this valid recovery action; scores are illustrative.}
\label{fig:motivation}
\end{figure}

\section{Introduction}

Outcome rewards indicate whether a multi-turn interaction succeeds but provide little guidance about which intermediate decisions should change. On-policy distillation (OPD) provides denser supervision while preserving student exploration by re-scoring student-generated responses~\cite{gkd2024,minillm2024}. Recent self-distillation methods use the same model as a synchronized teacher and condition its scoring on training-only feedback or references, such as successful trajectories or abstract task skills~\cite{sdpo2026,skillsd2026,sdar2026,serl2026}. The resulting teacher--student discrepancy provides a dense token-level signal.

When the reference is a successful trajectory, a direct approach supplies it to the teacher at every turn; we call this unconditional baseline \fullpath{}. However, each reference action is valid only from the state in which it was demonstrated, whereas the student's own actions change the execution state, such as its location, object properties, active page, or acquired tool results. After a detour or a reordered subgoal, the student can therefore reach a state from which a task-correct reference offers no compatible continuation. Figure~\ref{fig:motivation} illustrates this \emph{state--reference mismatch}: the reference reaches Product A from a results page, whereas the student is on Product B and samples the appropriate recovery action \texttt{Back to Search}. Conditioning the teacher on this reference can then lower its support for the student's valid action. Our fixed-state probes in Section~\ref{sec:teacher_probes} show this effect: with the state and student response held fixed, the successful reference raises teacher support for feasible actions at matched states but lowers it at mismatched states.

Prior work improves the reliability of multi-turn distillation by reshaping the rollout or prefix distribution on which the teacher is queried~\cite{tcod2026,reopd2026}, or by selecting, masking, or weighting the resulting teacher signal~\cite{sageopd2026,sdar2026,stepopsd2026,turnopd2026}. These approaches act before or after teacher scoring and leave the privileged context itself unchanged. State--reference mismatch, however, originates in that context: whether a reference is informative depends on how it relates to the state the student has reached. We therefore address reliability through the teacher context: whether a successful reference applies to the reached state and, if so, how to ground it in that state.

We introduce State-Matched Routing and Contextualized Self-Distillation (\method{}), which treats a successful trajectory as a state-indexed training resource. At each turn, it matches the reached state to a reference state and augments the reference with a reached-state summary and the compatible continuation for teacher scoring. Unmatched turns receive only GRPO supervision. State matching thus determines both when to use the reference and how to contextualize it, based on reference--state compatibility rather than teacher confidence, discrepancy, or outcomes.

To test whether this design generalizes beyond a single environment, we evaluate \method{} on three multi-turn agent settings with different forms of execution state: embodied interaction in ALFWorld~\cite{alfworld2021}, web navigation in WebShop~\cite{webshop2022}, and tool use in FTRL~\cite{ftrl2026}. Each instantiates the same steps of extracting task-relevant facts, checking whether a reference continuation applies, and contextualizing compatible references. Across Qwen2.5 and Qwen3 models from 1.7B to 4B, \method{} improves over unconditional \fullpath{}, and FTRL-trained policies transfer to BFCL~\cite{bfcl2025} and ToolHop~\cite{toolhop2025}.

Our contributions are threefold:
\begin{itemize}
    \item We identify \emph{state--reference mismatch} as an upstream reliability problem in privileged OPD: a task-correct executable reference may provide no state-aligned guidance for the state reached on policy.
    \item We introduce \method{}, which uses state--reference matching to jointly route reference-conditioned self-distillation and construct locally grounded teacher context.
    \item We show that a shared state-based design improves policies across embodied, web, and tool-use tasks and multiple model sizes, supported by transfer evaluation, probes, and ablations.
\end{itemize}

\section{Related Work}

\paragraph{Privileged on-policy and self-distillation.}
On-policy distillation trains the student on its own generated sequences, reducing the training--inference mismatch of classical distillation from fixed teacher outputs~\cite{hinton2015,kimrush2016,bornagain2018,minillm2024,gkd2024}. Privileged self-distillation removes the need for a separate teacher: the same model, conditioned on training-only traces or feedback, provides the target~\cite{opsdreasoner2026,sdpo2026}. Skill-SD and SDAR bring this approach to multi-turn agents through task skills and gated supervision~\cite{skillsd2026,sdar2026}, while SmartAD transfers trajectories from a larger tool-using teacher~\cite{smartad2026}. We instead study whether a given reference still applies once the student departs from it.

\paragraph{Reliable supervision in multi-turn interaction.}
Because earlier actions change later decision contexts, dense supervision must remain reliable as an interaction unfolds. One line of work controls the states on which the teacher is queried by reshaping rollout horizons or prefixes~\cite{tcod2026,reopd2026} or by accelerating interaction with reference-conditioned actions~\cite{actfirst2026}. Another regulates the resulting signal, selecting, weighting, or retiring it according to confidence, discrepancy, outcomes, or student progress~\cite{sageopd2026,sdar2026,stepopsd2026,hintsd2026,turnopd2026,retireopd2026}, or reshaping the distillation target~\cite{ocsd2026,turnsight2026,topd2026,aropd2026,dopd2026}. \method{} is complementary: it determines from reference--state compatibility whether the privileged context applies and adapts that context to the reached state.

\paragraph{State representations for multi-turn agents.}
In multi-turn interaction, what the next decision needs, namely what has been accomplished and what can still be done, is spread across a growing history. Making this information explicit as an intermediate state is therefore a common design. For inference, agents reason or reflect over the history~\cite{react2023,reflexion2023}, track goal states, progress, or compact summaries in its place~\cite{reflact2025,stateact2025,zipact2026,fromhistorytostate2026}, and organize memory, experience, or workflows by situation~\cite{samem2026,expel2024,synapse2024,awm2025}; explicit states also support the evaluation of tool-use agents~\cite{proxystate2026}. For training, states identify comparable decisions across rollouts: actions are grouped at repeated states for credit assignment~\cite{gigpo2025,hgpo2026,alignopsd2026}, and similar states anchor hindsight or retrieve references for distillation~\cite{serl2026,dartsd2026,gcopd2026}. We follow this practice but ask a different question of the state: whether the task's own successful reference still applies at the reached state and, if so, in what form the teacher should see it.

\section{Preliminaries}

\subsection{Multi-Turn On-Policy Learning}

Let $g$ denote an interactive task, given by an instruction and an environment instance. At turn $t$, the environment is in an execution state $s_t$; the agent observes $o_t$ and the admissible actions $\mathcal{A}_t$ and retains the history $h_t$ of earlier responses and observations. For tool use, $\mathcal{A}_t$ specifies the available tool schemas, and a response may invoke multiple tools. The ordinary prompt, sampled response, and parsed action are
\begin{align}
x_t&=\operatorname{Prompt}(g,h_t,o_t,\mathcal{A}_t),\\
y_t&\sim\pi_{\mathrm{old}}(\cdot\mid x_t),\qquad
a_t=\operatorname{Parse}(y_t).
\end{align}
Executing $a_t$ yields $s_{t+1}$ and $o_{t+1}$. An episode $\tau=(x_0,y_0,\ldots,x_T,y_T)$ runs until the task ends or the turn limit is reached and receives only a terminal reward $r(\tau)$. GRPO samples a group of episodes per task, normalizes their rewards within the group into trajectory-level advantages, and applies $\mathcal{L}_{\mathrm{GRPO}}$ to every episode~\cite{grpo2024}.

Privileged self-distillation adds dense supervision from a teacher $\pi_{\bar\theta}$, a gradient-detached copy of the current student $\pi_\theta$ that additionally receives training-only context $c_t$. After rollout, both score the same sampled response:
\begin{equation}
\pi_\theta(y_t\mid x_t), \qquad
\pi_{\bar\theta}(y_t\mid x_t,c_t).
\label{eq:student_teacher}
\end{equation}
The teacher generates no action, and $c_t$ is unavailable at inference. We compute the self-distillation loss (SDL) using the chosen-token K3 estimator of Skill-SD and SDAR~\cite{skillsd2026,sdar2026}. With binary route $w_t$ and turn-level term $\ell_{\mathrm{K3},t}(c_t)$, we optimize
\begin{equation}
\mathcal{L}=\mathcal{L}_{\mathrm{GRPO}}
+\lambda_{\mathrm{SDL}}
\frac{\sum_t w_t\ell_{\mathrm{K3},t}(c_t)}
{\sum_{t,i}m^{\mathrm{resp}}_{t,i}},
\label{eq:objective}
\end{equation}
where $m^{\mathrm{resp}}_{t,i}$ is the original response-token mask; routing affects only SDL.

\subsection{State-Conditional Successful References}

This paper studies privileged context built from a successful reference for the same task. Each training task $g$ has one reference path, taken from a benchmark walkthrough, the goal specification, or the dataset's tool calls (Appendix~\ref{sec:reference_construction}); evaluation uses no references. Omitting $g$ when the task is fixed, the path and its state-conditional transitions are
\begin{align}
p&=(\bar a_0,\ldots,\bar a_{n-1}),
\label{eq:reference_path}
\\
\bar s_k&\xrightarrow{\,\bar a_k\,}\bar s_{k+1},\qquad k<n,
\label{eq:reference_transition}
\end{align}
where $\bar s_k$ is the execution state before the $k$-th reference action. Success guarantees that $\bar a_k$ is valid from $\bar s_k$, not from an arbitrary state $s_t$ reached by the student. Since execution states are not directly exposed, we compare them through observable signatures $\sigma_t$ of $s_t$ and $\bar\sigma_k$ of $\bar s_k$ (Section~\ref{sec:state_signatures}). \fullpath{} sets $c_t=p$ at every turn; \method{} decides from $\sigma_t$ whether and how to use $p$.

\section{Why Privileged Guidance Must Be State-Compatible}

\subsection{State-Conditional Reference Validity}

After an on-policy detour or reordered subgoal, $p$ remains locally applicable only if some pre-action state $\bar s_k$ is compatible with the reached state $s_t$ and $\bar a_k$ can be grounded in $\mathcal{A}_t$. We say that $p$ \emph{supports} turn $t$ when such a position exists, and call the turn \emph{matched}; otherwise it is \emph{unmatched}. A matched turn identifies a position $k_t$ and its continuation $\bar a_{k_t}$.

We hypothesize that state--reference mismatch makes privileged guidance unreliable even when the reference solves the task. This raises two design questions: \emph{where} to apply distillation and \emph{how} to construct its context.

\begin{figure*}[!t]
\centering
\includegraphics[width=0.97\textwidth]{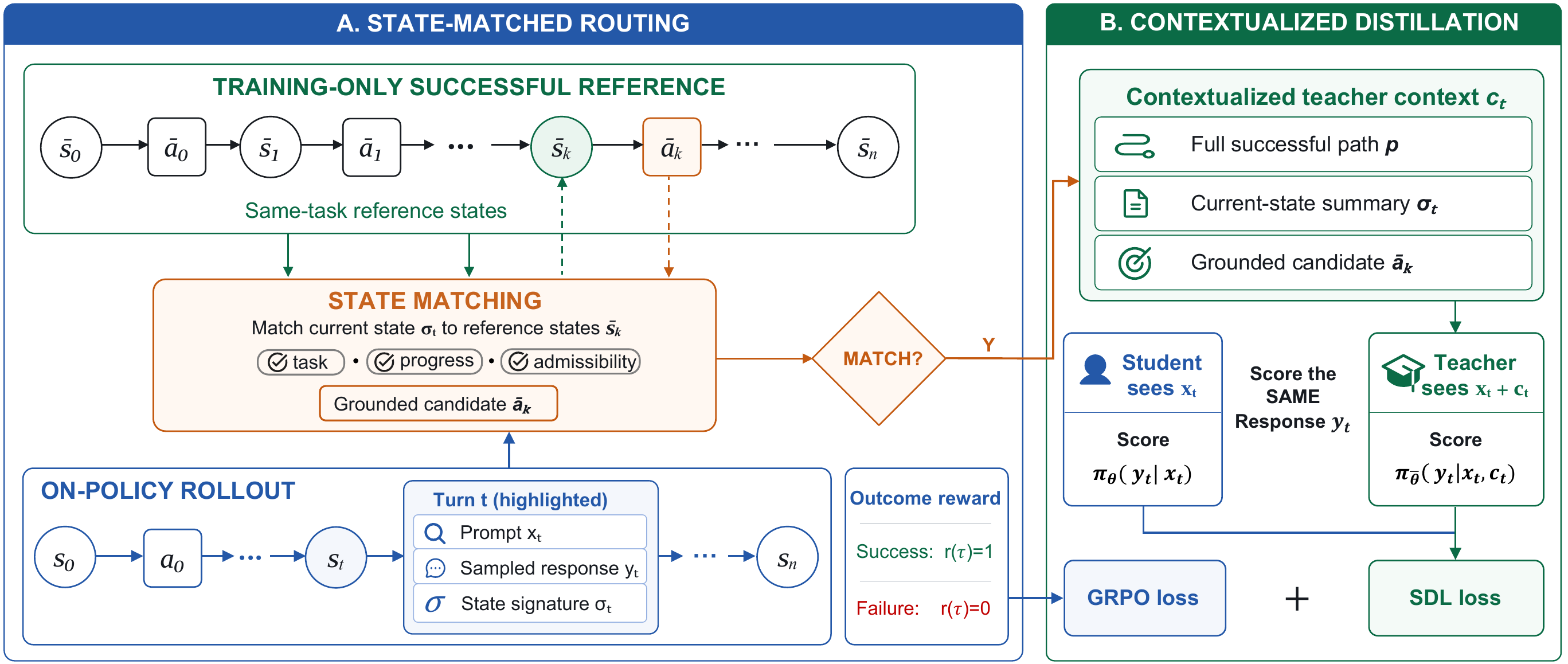}
\caption{\method{}: \textbf{A.} Match the reached state to a compatible reference transition. \textbf{B.} Condition teacher scoring on the path, current state, and matched candidate. GRPO trains all trajectories; only matched turns receive SDL.}
\label{fig:method}
\end{figure*}

\subsection{Fixed-State Teacher Probes}
\label{sec:teacher_probes}

The probes hold the reached state, ordinary prompt, and student response fixed and change only the teacher context. They address the two design questions in turn: does mismatch undermine guidance, and does contextualizing a compatible reference improve it? Each condition uses 400 ALFWorld turns. We average results from two frozen scorers; sampling details and 95\% confidence intervals appear in Appendix~\ref{sec:teacher_probe_details}.

\paragraph{Does mismatch undermine guidance?}
We compare feasible, progress-making student actions from successful trajectories in two conditions. At matched states, the action agrees with a continuation in the reference; at mismatched states, the reference provides no compatible continuation. Let $z$ denote the teacher context, $a_t$ the sampled action, and $\varnothing$ no privileged context. With the prompt and response fixed, we define the action score and its context-induced change:
\begin{align}
G_t(z)&=\overline{\log \pi_{\bar\theta}(a_t\mid x_t,z)},
\label{eq:teacher_action_score}\\
\Delta G_t(z)&=G_t(z)-G_t(\varnothing).
\label{eq:sampled_action_shift}
\end{align}
The overline averages over tokens in the action span. In this probe, $z$ is the task's full reference path. Positive $\Delta G_t$ indicates greater support for the feasible action, while negative $\Delta G_t$ indicates suppression.

The top block of Table~\ref{tab:teacher_audit} shows that the reference increases support at matched states ($+0.017$) but decreases it at mismatched states ($-0.052$), a difference of $+0.070$ (95\% CI $[+0.012,+0.153]$). A reference that solves the task can therefore work against a valid decision from the student's reached state, so state compatibility should be checked before privileged distillation is applied.

\begin{table}[!t]
\centering
\begingroup
\fontsize{10}{12}\selectfont
\setlength{\tabcolsep}{1.5mm}
\begin{tabular*}{\columnwidth}{@{\extracolsep{\fill}}lr@{}}
\toprule
State alignment & Score change $\Delta G$ \\
\midrule
\multicolumn{2}{l}{{\fontsize{9}{11}\selectfont\textit{Feasible actions from successful trajectories}}} \\
Matched & $+0.017$ \\
Mismatched & $-0.052$ \\
\midrule
Teacher context & Margin change $\Delta M$ \\
\midrule
\multicolumn{2}{l}{{\fontsize{9}{11}\selectfont\textit{Matched states; sampled action differs from reference}}} \\
No privilege & $0.000$ \\
Abstract skill & $+0.153$ \\
Random successful path & $+0.442$ \\
Reference path & $+1.158$ \\
Contextualized reference path & $\mathbf{+1.424}$ \\
\bottomrule
\end{tabular*}
\endgroup
\caption{Fixed-state teacher probes on ALFWorld. $\Delta G$ measures support for feasible actions; $\Delta M$ measures preference for the reference action. Both are changes from no privilege. 95\% CIs are in Appendix~\ref{sec:teacher_probe_details}.}
\label{tab:teacher_audit}
\end{table}

\paragraph{Does state context improve compatible guidance?}
Conditioned on a match, we compare context designs on separate matched turns where sampled and reference actions differ. We replace only the sampled action span with the compatible reference action $\bar a_{k_t}$, retaining the original reasoning and all other response tokens, and define the reference-minus-sampled action margin and its change from no privileged context:
\begin{align}
M_t(z)&=\overline{\log \pi_{\bar\theta}(\bar a_{k_t}\mid x_t,z)}-G_t(z),
\label{eq:teacher_margin}\\
\Delta M_t(z)&=M_t(z)-M_t(\varnothing).
\label{eq:teacher_margin_shift}
\end{align}
Positive $\Delta M_t$ indicates a shift toward the reference action relative to the sampled action. The bottom block of Table~\ref{tab:teacher_audit} shows that an abstract skill or a random successful path from another task in the same family provides less guidance than the task's own reference path. Contextualizing this reference path with the reached-state summary and compatible action raises the preference shift from $+1.158$ to $+1.424$. The reference is thus more useful when explicitly connected to the current state.

\section{State-Matched Routing and Contextualized Self-Distillation}

The probes suggest two controls: withhold reference-conditioned SDL when no compatible transition exists, and ground the guidance in the reached state when one does. As Figure~\ref{fig:method} shows, state-matched routing determines \emph{where} to apply SDL, and contextualized guidance determines \emph{what} the teacher receives.

\subsection{Reconstructing State Signatures}
\label{sec:state_signatures}

A signature records the facts on which reference actions depend, is maintained from observable interaction, and is computed in the same form for reference prefixes. It should preserve distinctions that change which next action is valid, while allowing different histories to match when they reach compatible states. As in agents that track explicit task state (Section~2), these facts come from the environment's own structure: they are the preconditions and effects of reference actions, which the action semantics and the successful reference itself expose. ALFWorld tracks location, inventory, and task-object locations and properties, updated by navigation, placement, and transformation actions; WebShop tracks page type, current product, and selected options; FTRL represents the state as the set of observed tool calls and their returned values (Appendix~\ref{sec:state_adapters}). No component is learned; on ALFWorld, signature construction, matching, and context rendering take $0.83$\,ms of single-thread CPU time per matched turn (Table~\ref{tab:supp_cpu_runtime}).

The signatures of the reference state $\bar s_k$ and the reached state $s_t$ are
\begin{equation}
\begin{aligned}
\bar{\sigma}_k&=\phi^{\mathrm{ref}}(g,\bar a_{<k},\bar o_{\le k}),\\
\sigma_t&=\phi^{\mathrm{stu}}(g,h_t,o_t,\mathcal{A}_t),
\end{aligned}
\label{eq:state_signatures}
\end{equation}
where $\bar o_{\le k}$ denotes observations available before $\bar a_k$. Environment-specific fields and construction rules appear in Appendix~\ref{sec:state_adapters}.

\begin{table*}[!t]
\centering
\begingroup
\fontsize{10}{11}\selectfont
\setlength{\tabcolsep}{2pt}
\begin{tabular*}{0.96\textwidth}{@{\extracolsep{\fill}}l*{9}{>{\centering\arraybackslash}p{0.074\textwidth}}@{}}
\toprule
& \multicolumn{7}{c}{ALFWorld} & \multicolumn{2}{c}{WebShop} \\
\cmidrule(lr){2-8}\cmidrule(lr){9-10}
Method & Pick & Look & Clean & Heat & Cool & Pick2 & Average & Score & Acc \\
\midrule
\rowcolor{gray!15}
\multicolumn{10}{l}{\textbf{Qwen2.5-3B-Instruct}} \\
Vanilla & $41.1$ & $28.1$ & $2.9$ & $6.2$ & $3.1$ & $7.2$ & $16.4$ & $7.2$ & $0.6$ \\
GRPO & $91.2$ & $\underline{62.5}$ & $\underline{96.2}$ & $61.9$ & $65.0$ & $47.4$ & $75.0$ & $79.8$ & $63.3$ \\
Skill-SD & $88.2$ & $50.0$ & $\underline{96.2}$ & $52.4$ & $65.0$ & $57.9$ & $73.4$ & $75.9$ & $64.0$ \\
SDAR & $\underline{97.1}$ & $\underline{62.5}$ & $\mathbf{100.0}$ & $61.9$ & $75.0$ & $\mathbf{84.2}$ & $\underline{84.4}$ & $\underline{85.0}$ & $68.0$ \\
\fullpath{} & $93.5$ & $46.9$ & $81.0$ & $\mathbf{88.7}$ & $\underline{83.3}$ & $43.8$ & $76.6$ & $84.2$ & $\underline{73.4}$ \\
\midrule
\method{} (ours) & $\mathbf{98.7}$ & $\mathbf{76.6}$ & $89.8$ & $\underline{82.5}$ & $\mathbf{89.2}$ & $\underline{81.0}$ & $\mathbf{88.3}$ & $\mathbf{86.3}$ & $\mathbf{73.6}$ \\
\midrule
\rowcolor{gray!15}
\multicolumn{10}{l}{\textbf{Qwen3-1.7B}} \\
Vanilla & $9.1$ & $29.7$ & $3.6$ & $0.0$ & $0.0$ & $3.8$ & $6.8$ & $47.3$ & $2.7$ \\
GRPO & $\underline{88.4}$ & $56.2$ & $69.7$ & $70.0$ & $71.2$ & $47.5$ & $71.7$ & $67.3$ & $38.3$ \\
Skill-SD & $47.2$ & $37.5$ & $47.5$ & $42.5$ & $29.9$ & $2.9$ & $37.9$ & $\underline{81.8}$ & $53.9$ \\
SDAR & $63.9$ & $62.5$ & $72.0$ & $43.8$ & $63.2$ & $37.5$ & $57.8$ & $76.8$ & $\underline{58.6}$ \\
OCSD & $81.2$ & $52.1$ & $71.2$ & $68.3$ & $44.0$ & $\underline{50.0}$ & $63.5$ & $74.4$ & $50.6$ \\
\fullpath{} & $82.0$ & $\underline{73.4}$ & $\underline{73.9}$ & $\mathbf{88.7}$ & $\underline{79.5}$ & $42.5$ & $\underline{74.6}$ & $69.4$ & $57.4$ \\
\midrule
\method{} (ours) & $\mathbf{95.4}$ & $\mathbf{76.6}$ & $\mathbf{93.5}$ & $\underline{87.5}$ & $\mathbf{87.5}$ & $\mathbf{65.6}$ & $\mathbf{86.5}$ & $\mathbf{82.5}$ & $\mathbf{69.3}$ \\
\midrule
\rowcolor{gray!15}
\multicolumn{10}{l}{\textbf{Qwen3-4B}} \\
Vanilla & $57.8$ & $32.6$ & $15.4$ & $20.6$ & $7.9$ & $20.1$ & $27.8$ & $14.7$ & $1.6$ \\
GRPO & $86.6$ & $59.5$ & $77.7$ & $67.1$ & $55.5$ & $55.0$ & $70.6$ & $81.9$ & $69.5$ \\
OPSD & $69.9$ & $39.2$ & $26.6$ & $26.6$ & $19.0$ & $26.4$ & $36.2$ & $27.8$ & $4.7$ \\
GRPO+OPSD & $88.0$ & $52.8$ & $75.0$ & $62.5$ & $68.3$ & $36.9$ & $68.8$ & $82.5$ & $71.3$ \\
RLSD & $87.9$ & $\mathbf{83.4}$ & $80.5$ & $70.3$ & $79.6$ & $60.6$ & $79.2$ & $83.7$ & $68.5$ \\
SDAR & $85.2$ & $63.9$ & $76.8$ & $80.0$ & $72.9$ & $42.6$ & $73.2$ & $83.6$ & $71.9$ \\
OCSD & $\mathbf{91.9}$ & $\underline{77.2}$ & $\underline{90.2}$ & $84.7$ & $\underline{80.9}$ & $\underline{63.8}$ & $\underline{82.8}$ & $\mathbf{86.9}$ & $\underline{73.7}$ \\
\fullpath{} & $89.3$ & $76.4$ & $\mathbf{90.4}$ & $\underline{87.8}$ & $79.4$ & $52.7$ & $80.2$ & $82.4$ & $73.6$ \\
\midrule
\method{} (ours) & $\underline{91.4}$ & $76.4$ & $88.5$ & $\mathbf{90.0}$ & $\mathbf{85.3}$ & $\mathbf{65.5}$ & $\mathbf{83.8}$ & $\underline{86.0}$ & $\mathbf{74.3}$ \\
\bottomrule
\end{tabular*}
\endgroup
\caption{ALFWorld and WebShop performance (\%). Bold/underline: best/second-best within each model. Following OCSD~\cite{ocsd2026}, Qwen3-4B results are averaged over three seeds; standard deviations are in Appendix~\ref{sec:multiseed_results}}
\label{tab:main}
\end{table*}

\begin{table*}[tp]
\centering
\begingroup
\fontsize{10}{12}\selectfont
\newcommand{\metricheader}[1]{{\fontsize{9}{11}\selectfont #1}}
\setlength{\tabcolsep}{0.7mm}
\begin{tabular*}{\textwidth}{@{\extracolsep{\fill}}l*{9}{r}cc@{}}
\toprule
\multirow{3}{*}{Method} & \multicolumn{3}{c}{FTRL (ID)} & \multicolumn{6}{c}{BFCL} & \multicolumn{1}{c}{ToolHop} & \multirow{3}{*}{Overall} \\
\cmidrule(lr){2-4}\cmidrule(lr){5-10}\cmidrule(lr){11-11}
& \multirow{2}{*}{\metricheader{Solve-P}} & \multirow{2}{*}{\metricheader{Solve-R}} & \multirow{2}{*}{\metricheader{Solve-F1}} & \multicolumn{4}{c}{\metricheader{Multi-Turn}} & \multicolumn{2}{c}{\metricheader{Agentic}} & \multicolumn{1}{c}{\multirow{2}{*}{\metricheader{Acc.}}} & \\
\cmidrule(lr){5-8}\cmidrule(lr){9-10}
& & & & \multicolumn{1}{c}{\metricheader{Base}} & \multicolumn{1}{c}{\metricheader{MF}} & \multicolumn{1}{c}{\metricheader{MP}} & \multicolumn{1}{c}{\metricheader{LC}} & \multicolumn{1}{c}{\metricheader{Search}} & \multicolumn{1}{c}{\metricheader{Memory}} & & \\
\midrule
Vanilla & $17.84$ & $24.45$ & $18.50$ & $38.00$ & $37.50$ & $28.00$ & $26.00$ & $12.00$ & $21.72$ & $25.93$ & $23.60$ \\
GRPO & $31.61$ & $32.64$ & $28.49$ & $42.00$ & $34.00$ & $29.00$ & $23.50$ & $8.00$ & $19.35$ & $21.98$ & $25.26$ \\
ToolRL & $34.07$ & $31.63$ & $27.94$ & $43.00$ & $41.00$ & $27.50$ & $25.50$ & $8.50$ & $\underline{24.09}$ & $23.65$ & $26.71$ \\
MatchTIR & $37.77$ & $36.44$ & $33.51$ & $51.00$ & $44.00$ & $30.50$ & $38.00$ & $\underline{18.50}$ & $23.01$ & $37.56$ & $34.76$ \\
SDPO & $31.35$ & $32.66$ & $27.71$ & $43.50$ & $41.50$ & $31.00$ & $29.50$ & $8.00$ & $22.15$ & $26.53$ & $27.61$ \\
RLSD & $34.19$ & $32.63$ & $31.13$ & $45.50$ & $39.00$ & $33.50$ & $31.00$ & $12.00$ & $21.93$ & $33.30$ & $31.02$ \\
SDAR & $33.67$ & $37.03$ & $31.95$ & $47.00$ & $40.50$ & $30.50$ & $37.00$ & $16.00$ & $22.80$ & $37.36$ & $33.55$ \\
SOD & $31.89$ & $34.91$ & $30.66$ & $38.50$ & $34.00$ & $28.00$ & $24.50$ & $9.50$ & $21.72$ & $22.28$ & $26.06$ \\
TurnSight & $\underline{44.61}$ & $40.16$ & $38.44$ & $50.50$ & $\underline{46.50}$ & $\underline{36.00}$ & $\underline{41.00}$ & $\mathbf{21.00}$ & $\mathbf{24.73}$ & $38.29$ & $37.51$ \\
\fullpath{} & $\mathbf{45.45}$ & $\underline{43.57}$ & $\underline{43.56}$ & $\underline{54.50}$ & $\mathbf{48.00}$ & $31.00$ & $\mathbf{43.00}$ & $14.50$ & $20.65$ & $\underline{44.72}$ & $\underline{39.92}$ \\
\midrule
\method{} (ours) & $43.70$ & $\mathbf{46.36}$ & $\mathbf{43.99}$ & $\mathbf{58.00}$ & $46.00$ & $\mathbf{37.50}$ & $37.50$ & $17.00$ & $21.72$ & $\mathbf{47.44}$ & $\mathbf{41.39}$ \\
\bottomrule
\end{tabular*}
\endgroup
\caption{Qwen3-4B tool-use performance (\%). BFCL and ToolHop are out of distribution. MF/MP/LC: Miss Function/Miss Parameter/Long Context. Bold/underline: best/second-best.}
\label{tab:ftrl}
\end{table*}

\subsection{Structured-State Matching and Routing}
\label{sec:matching}

The matcher retrieves the reference of the same task. Grounding $\Gamma(\bar a_k;\mathcal{A}_t)$ maps a reference action to its admissible form at turn $t$ and returns $\varnothing$ if none exists; for order-invariant subgoals, it may return a set of compatible actions. Let $\sigma_t\models\bar\sigma_k$ denote that the current signature satisfies the task-relevant requirements of position $k$. Position $k$ is compatible when
\begin{equation}
C_t(k)=
\mathbb{I}[\sigma_t\models\bar{\sigma}_k]\,
\mathbb{I}[\Gamma(\bar a_k;\mathcal{A}_t)\neq\varnothing].
\label{eq:compatibility}
\end{equation}
Together with same-task retrieval, $C_t(k)$ enforces task identity, state compatibility, and action admissibility; environment-specific rules appear in Appendix~\ref{sec:state_adapters}. \method{} selects the latest compatible position and routes SDL iff one exists:
\begin{equation}
\begin{aligned}
k_t&=\max\{k:C_t(k)=1\},\\
w_t&=\mathbb{I}\!\left[\textstyle\max_k C_t(k)=1\right].
\end{aligned}
\label{eq:routing}
\end{equation}
The latest position avoids repeating completed subgoals. On a matched turn, the candidate is the grounded action $\Gamma(\bar a_{k_t};\mathcal{A}_t)$, which we write as $\bar a_{k_t}$ following Figure~\ref{fig:method}. Unmatched turns receive only GRPO.

\subsection{State-Contextualized Guidance Construction}

For a matched turn, \method{} renders
\begin{equation}
c_t^{\mathrm{SMRC}}=\operatorname{Render}\!\left(
p,\sigma_t,\bar a_{k_t}
\right).
\label{eq:guidance_context}
\end{equation}
The full path provides the task-level plan, the current-state summary rendered from $\sigma_t$ locates reached progress, and the grounded candidate identifies a compatible next transition. Stating this relation explicitly spares the teacher from inferring it from the path alone; prompt details are in Appendix~\ref{sec:state_adapters}.

Substituting $w_t$ and $c_t^{\mathrm{SMRC}}$ into Equation~\ref{eq:objective} applies GRPO+SDL on matched turns and GRPO alone otherwise. At inference, the policy uses only the ordinary prompt $x_t$, without references, signatures, matching, or teacher context.

\section{Experimental Setup}

\paragraph{Datasets and metrics.}
ALFWorld~\cite{alfworld2021} instantiates ALFRED household tasks~\cite{alfred2020} in TextWorld~\cite{textworld2018}. WebShop~\cite{webshop2022} requires agents to search for and purchase products satisfying user requirements. We evaluate on 128 fixed validation tasks per environment with four rollouts each. ALFWorld Average and WebShop Acc measure task-averaged success (Avg@4); WebShop Score is the graded task reward.
FTRL~\cite{ftrl2026} covers single-hop, parallel, and multi-hop tool use, with 2,215 training and 200 evaluation tasks. We report Solve-P/R/F1, which measure acquisition of the required tool information. We further evaluate the FTRL-trained policies on BFCL~\cite{bfcl2025} Multi-Turn, Search, and Memory, and ToolHop~\cite{toolhop2025}, without additional training.

\paragraph{Baselines and comparisons.}
Vanilla is the base model and GRPO~\cite{grpo2024} uses terminal rewards only. Self-distillation baselines are OPSD~\cite{opsdreasoner2026}, GRPO+OPSD, RLSD~\cite{rlsd2026}, Skill-SD~\cite{skillsd2026}, SDAR~\cite{sdar2026}, and OCSD~\cite{ocsd2026}; FTRL adds ToolRL~\cite{toolrl2025}, MatchTIR~\cite{matchtir2026}, SDPO~\cite{sdpo2026}, SOD~\cite{sod2026}, and TurnSight~\cite{turnsight2026}. Our primary baseline, \fullpath{}, supplies the full reference at every turn. Appendix~\ref{sec:4b_protocols} lists result sources.

\paragraph{Implementation details.}
We use Qwen2.5-3B-Instruct~\cite{qwen252024} and Qwen3-1.7B/4B~\cite{qwen32025}. Following prior work~\cite{sdar2026,ocsd2026}, all methods share the same optimization settings, including learning rate $10^{-6}$, eight rollouts per task, and $\lambda_{\mathrm{SDL}}=0.01$. \fullpath{} and \method{} differ only in the route $w_t$ and the teacher context $c_t$. Qwen3-1.7B is trained for 250 updates on ALFWorld, other ALFWorld and WebShop runs for 150, and FTRL runs for 207 updates (three epochs) following TurnSight~\cite{turnsight2026}. We report final-checkpoint results; full settings are in Appendix~\ref{sec:hyperparameters}.

\section{Main Results}

\subsection{ALFWorld and WebShop}

Table~\ref{tab:main} shows gains over \fullpath{} across all three model sizes and both environments. With Qwen3-1.7B, \method{} raises ALFWorld Average from $74.6\%$ to $86.5\%$, WebShop Score from $69.4\%$ to $82.5\%$, and Acc from $57.4\%$ to $69.3\%$. ALFWorld gains are largest on Clean and Pick Two and remain positive on most task families. With Qwen2.5-3B, ALFWorld Average rises from $76.6\%$ to $88.3\%$, with gains on WebShop as well. With Qwen3-4B, ALFWorld Average improves from $80.2\%$ to $83.8\%$ and WebShop Score from $82.4\%$ to $86.0\%$; \method{} also achieves the best ALFWorld Average and WebShop Acc among all 4B methods, including OCSD.

\begin{figure}[!t]
\centering
\includegraphics[width=0.9\columnwidth]{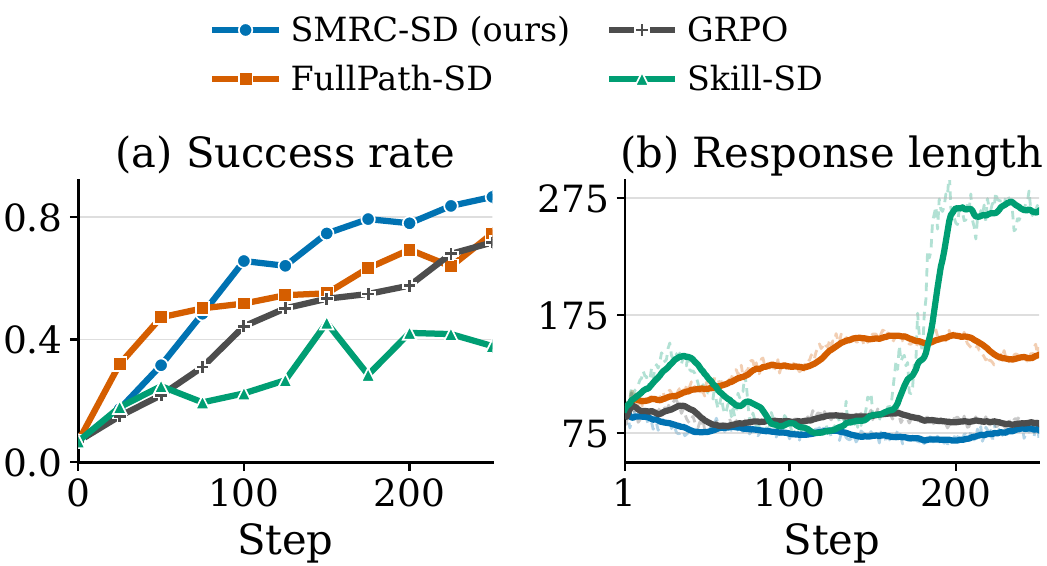}
\caption{Qwen3-1.7B ALFWorld training: (a) Avg@4 evaluated every 25 updates; (b) response length, with per-update values dashed and 15-update means solid.}
\label{fig:training_curves}
\end{figure}

Figure~\ref{fig:training_curves} shows that Qwen3-1.7B has not converged by update 150 across methods, motivating a shared 250-update budget. At update 150, \method{} already achieves $74.6\%$ Avg@4 versus $55.1\%$ for \fullpath{} and remains ahead thereafter. Its responses stay close to GRPO in length and are shorter and less repetitive than the distillation baselines at the last checkpoint (Appendix~\ref{sec:response_style}).

\subsection{Tool Use and Policy Transfer}
Table~\ref{tab:ftrl} extends the approach to tool use. Over \fullpath{}, \method{} improves FTRL Solve-F1 from $43.56\%$ to $43.99\%$, and the gains are larger out of distribution: BFCL Multi-Turn Base rises from $54.50\%$ to $58.00\%$, ToolHop accuracy from $44.72\%$ to $47.44\%$, and Overall from $39.92\%$ to $41.39\%$, the best among all methods. Observed tool results thus provide a useful state representation for training, and the resulting policies transfer to other tool-use benchmarks without the teacher or matcher.

\begin{table}
\centering
\begingroup
\setlength{\tabcolsep}{1mm}
\begingroup
\fontsize{10}{12}\selectfont
\begin{tabular}{@{}P{0.29\columnwidth}P{0.25\columnwidth}P{0.22\columnwidth}r@{}}
\toprule
Variant & Matched & Unmatched & Avg@4 \\
\midrule
\fullpath{} & Full path & Full path & $74.6$ \\
+ Routing & Full path & None & $83.6$ \\
+ Dynamic Context & SMRC context & Full path & $69.5$ \\
\method{} & SMRC context & None & $\mathbf{86.5}$ \\
\bottomrule
\end{tabular}
\endgroup
\endgroup
\caption{Guidance routing and construction ablation on ALFWorld (Avg@4, \%).}
\label{tab:guidance_routing}
\end{table}

\section{Analysis and Ablations}

\subsection{Routing and Context Construction}

All ablations use Qwen3-1.7B on ALFWorld. Table~\ref{tab:guidance_routing} adds routing and dynamic context to \fullpath{} separately and jointly.

Routing alone raises Avg@4 from $74.6\%$ to $83.6\%$, and adding SMRC context raises it further to $86.5\%$. Without routing, SMRC context on matched turns does not help: plain full-path SDL on the remaining unmatched turns lowers Avg@4 to $69.5\%$. Mismatched guidance therefore dominates when it is left in place, and contextualized guidance pays off once routing removes it.

\subsection{Which Turns Are Distilled}

Routing also reduces the number of distilled turns. To separate which turns are distilled from how many, Table~\ref{tab:routing_selection} fixes the teacher context to FullPath and varies the selected turns. At each update, the random control selects as many turns as the matcher identifies in its own batch; run-level ratios differ only because the trajectories differ.

\begin{table}[!ht]
\centering
\begingroup
\setlength{\tabcolsep}{1mm}
\begingroup
\fontsize{10}{12}\selectfont
\begin{tabular}{@{}P{0.45\columnwidth}cr@{}}
\toprule
FullPath-SD applied to & \shortstack{Selected (\%)\\mean (final)} & Avg@4 \\
\midrule
All turns & $100.0\;(100.0)$ & $74.6$ \\
Matched turns & $15.3\;(31.8)$ & $\mathbf{83.6}$ \\
Random turns (same count) & $16.5\;(32.1)$ & $72.3$ \\
Unmatched turns (all) & $82.1\;(73.6)$ & $75.0$ \\
\bottomrule
\end{tabular}
\endgroup
\endgroup
\caption{\fullpath{} turn-selection controls on ALFWorld (\%).}
\label{tab:routing_selection}
\end{table}

Matched selection outperforms the same-count random control by $11.3$ points, although the random control distills slightly more response tokens ($16.4\%$ vs.\ $14.2\%$; Appendix~\ref{sec:matcher_audit}), so the gain does not come from weaker distillation. Distilling only unmatched turns reaches $75.0\%$, close to all-turn \fullpath{}, as these turns dominate coverage.

\subsection{Teacher-Context Components}

\begin{table}[!ht]
\centering
\begingroup
\setlength{\tabcolsep}{1mm}
\begingroup
\fontsize{10}{12}\selectfont
\begin{tabular}{@{}P{0.43\columnwidth}cccr@{}}
\toprule
Context on matched turns & Path & State & Cand. & Avg@4 \\
\midrule
FullPath & $\checkmark$ & -- & -- & $83.6$ \\
FullPath + candidate & $\checkmark$ & -- & $\checkmark$ & $82.0$ \\
FullPath + state summary & $\checkmark$ & $\checkmark$ & -- & $83.4$ \\
SMRC context & $\checkmark$ & $\checkmark$ & $\checkmark$ & $\mathbf{86.5}$ \\
\bottomrule
\end{tabular}
\endgroup
\endgroup
\caption{Teacher-context component ablation on matched ALFWorld turns (Avg@4, \%).}
\label{tab:guidance_components}
\end{table}

With routing fixed (Table~\ref{tab:guidance_components}), neither the candidate ($82.0\%$) nor the state summary ($83.4\%$) alone improves over FullPath ($83.6\%$); their combination reaches $86.5\%$. The summary locates reached progress and the candidate specifies the continuation from it, so the teacher needs both to connect the current state to the full path.

\subsection{Matcher Coverage and Executability}

Table~\ref{tab:matcher} compares structured-state matching with a history matcher based on observation and action-history similarity.

\begin{table}[!ht]
\centering
\begingroup
\setlength{\tabcolsep}{1mm}
\begingroup
\fontsize{10}{12}\selectfont
\begin{tabular}{lrr}
\toprule
Metric & \shortstack{History\\matcher} & \shortstack{Structured\\state} \\
\midrule
Candidate coverage & $15.4\%$ & $\mathbf{20.2\%}$ \\
Shared matches & $98.8\%$ & $75.4\%$ \\
Avg@4 & $75.6\%$ & $\mathbf{86.5\%}$ \\
Replay success & $99.0\%$ & $\mathbf{100\%}$ \\
\bottomrule
\end{tabular}
\endgroup
\endgroup
\caption{ALFWorld matcher comparison. Replay uses independent samples. Shared matches: intersection as a fraction of each matcher's matches.}
\label{tab:matcher}
\end{table}

On 35,712 archived turns, structured matching expands coverage from $15.4\%$ to $20.2\%$ while retaining $98.8\%$ of history matches. Most history-only matches are rejected because the inventory no longer supports the continuation. Structured state thus recognizes progress reached through different histories while checking execution constraints; it also improves Avg@4 from $75.6\%$ to $86.5\%$.

Replaying each student's action prefix, followed by the selected candidate and reference suffix, succeeds for $792/800$ history matches and $781/781$ independently sampled structured-state matches (Appendix~\ref{sec:matcher_audit}).

\section{Conclusion}

A successful reference guides the student only from states it supports; under mismatch, it can undermine guidance. \method{} uses state compatibility to decide when to distill and how to construct teacher context, improving policies across embodied, web, and tool-use tasks. Routing removes harmful guidance, and contextualization strengthens the guidance that remains. More broadly, our results suggest treating privileged context as a function of the reached state rather than a fixed property.

\pagebreak
\clearpage
\section*{Limitations}

\method{} uses training-time references and a task-relevant state representation. In our three environments, state fields and compatibility rules follow directly from action preconditions and effects. In settings where observations are mainly raw pixels or free-form dialogue, exact compatibility rules are harder to specify, and a learned or model-extracted state with approximate matching could take their place. We also evaluate one canonical reference per task. When several successful references are available, the matcher could search the states of all of them, but we have not tested how this affects coverage or the choice of guidance.

\bibliography{references}

@misc{ocsd2026,
  title = {Agentic Reinforcement Learning with Observation-Calibrated Self-Distillation},
  author = {Yang, Yi and Qin, Cong and Liu, Xiaodan and Chen, Chishui and Dong, Qing and Zhang, Yan and Liu, Cao and Yang, Zhao and Pan, Lu and Lin, Jiaye and Feng, Yi},
  year = {2026},
  eprint = {2608.04788},
  archiveprefix = {arXiv},
  url = {https://arxiv.org/abs/2608.04788v1}
}

@misc{turnsight2026,
  title = {{TurnSight}: Turn-Level Hindsight Self-Distillation for Tool-Integrated Reasoning},
  author = {Qu, Changle and Dai, Sunhao and Cai, Hengyi and Zhou, Yuqi and Chen, Xinran and Simon and Xu, Jun},
  year = {2026},
  eprint = {2608.04007},
  archiveprefix = {arXiv},
  url = {https://arxiv.org/abs/2608.04007v1}
}

@misc{hinton2015,
  title = {Distilling the Knowledge in a Neural Network},
  author = {Hinton, Geoffrey and Vinyals, Oriol and Dean, Jeff},
  year = {2015},
  eprint = {1503.02531},
  archiveprefix = {arXiv},
  url = {https://arxiv.org/abs/1503.02531}
}

@inproceedings{kimrush2016,
  title = {Sequence-Level Knowledge Distillation},
  author = {Kim, Yoon and Rush, Alexander M.},
  editor = {Su, Jian and Duh, Kevin and Carreras, Xavier},
  booktitle = {Proceedings of the 2016 Conference on Empirical Methods in Natural Language Processing},
  month = {nov},
  year = {2016},
  address = {Austin, Texas},
  publisher = {Association for Computational Linguistics},
  url = {https://aclanthology.org/D16-1139/},
  doi = {10.18653/v1/D16-1139},
  pages = {1317--1327}
}

@inproceedings{bornagain2018,
  title = {Born Again Neural Networks},
  author = {Furlanello, Tommaso and Lipton, Zachary and Tschannen, Michael and Itti, Laurent and Anandkumar, Anima},
  booktitle = {Proceedings of the 35th International Conference on Machine Learning},
  year = {2018},
  volume = {80},
  series = {Proceedings of Machine Learning Research},
  pages = {1607--1616},
  publisher = {PMLR},
  url = {https://proceedings.mlr.press/v80/furlanello18a.html}
}

@inproceedings{gkd2024,
  author = {Agarwal, Rishabh and Vieillard, Nino and Zhou, Yongchao and Stanczyk, Piotr and Ramos Garea, Sabela and Geist, Matthieu and Bachem, Olivier},
  booktitle = {International Conference on Learning Representations},
  editor = {B. Kim and Y. Yue and S. Chaudhuri and K. Fragkiadaki and M. Khan and Y. Sun},
  pages = {21246--21263},
  title = {On-Policy Distillation of Language Models: Learning from Self-Generated Mistakes},
  url = {https://proceedings.iclr.cc/paper_files/paper/2024/file/5be69a584901a26c521c2b51e40a4c20-Paper-Conference.pdf},
  volume = {2024},
  year = {2024}
}

@inproceedings{minillm2024,
  author = {Gu, Yuxian and Dong, Li and Wei, Furu and Huang, Minlie},
  booktitle = {International Conference on Learning Representations},
  editor = {B. Kim and Y. Yue and S. Chaudhuri and K. Fragkiadaki and M. Khan and Y. Sun},
  pages = {32694--32717},
  title = {{MiniLLM}: Knowledge Distillation of Large Language Models},
  url = {https://proceedings.iclr.cc/paper_files/paper/2024/file/8ac015d409635f196f9e3e9dcfb9a94e-Paper-Conference.pdf},
  volume = {2024},
  year = {2024}
}

@inproceedings{sdpo2026,
  title = {Reinforcement Learning via Self-Distillation},
  author = {H\"{u}botter, Jonas and L\"{u}beck, Frederike and Behric, Lejs Deen and Baumann, Anton and Bagatella, Marco and Marta, Daniel and Hakimi, Ido and Shenfeld, Idan and Kleine Buening, Thomas and Guestrin, Carlos and Krause, Andreas},
  booktitle = {Proceedings of the 43rd International Conference on Machine Learning},
  pages = {48677--48721},
  year = {2026},
  editor = {Zhang, Tong and Dudik, Miroslav and Jaggi, Martin and Agarwal, Alekh and Li, Sharon and Schuurmans, Dale and Zhu, Jerry and Berkenkamp, Felix and Dong, Hanze and Bietti, Alberto},
  volume = {306},
  series = {Proceedings of Machine Learning Research},
  month = {06--11 Jul},
  publisher = {PMLR},
  url = {https://proceedings.mlr.press/v306/hubotter26a.html}
}

@inproceedings{opsdreasoner2026,
  title = {Self-Distilled Reasoner: On-Policy Self-Distillation for Large Language Models},
  author = {Zhao, Siyan and Xie, Zhihui and Liu, Mengchen and Huang, Jing and Pang, Guan and Chen, Feiyu and Grover, Aditya},
  booktitle = {Proceedings of the 43rd International Conference on Machine Learning},
  pages = {162433--162448},
  year = {2026},
  editor = {Zhang, Tong and Dudik, Miroslav and Jaggi, Martin and Agarwal, Alekh and Li, Sharon and Schuurmans, Dale and Zhu, Jerry and Berkenkamp, Felix and Dong, Hanze and Bietti, Alberto},
  volume = {306},
  series = {Proceedings of Machine Learning Research},
  month = {06--11 Jul},
  publisher = {PMLR},
  url = {https://proceedings.mlr.press/v306/zhao26be.html}
}

@misc{skillsd2026,
  title = {{Skill-SD}: Skill-Conditioned Self-Distillation for Multi-turn {LLM} Agents},
  author = {Wang, Hao and Wang, Guozhi and Xiao, Han and Zhou, Yufeng and Pan, Yue and Wang, Jichao and Xu, Ke and Wen, Yafei and Ruan, Xiaohu and Chen, Xiaoxin and Qi, Honggang},
  year = {2026},
  eprint = {2604.10674},
  archiveprefix = {arXiv},
  url = {https://arxiv.org/abs/2604.10674}
}

@misc{sdar2026,
  title = {Self-Distilled Agentic Reinforcement Learning},
  author = {Lu, Zhengxi and Yao, Zhiyuan and Han, Zhuowen and Wang, Zi-Han and Wu, Jinyang and Gu, Qi and Cai, Xunliang and Lu, Weiming and Xiao, Jun and Zhuang, Yueting and Shen, Yongliang},
  year = {2026},
  eprint = {2605.15155},
  archiveprefix = {arXiv},
  url = {https://arxiv.org/abs/2605.15155}
}

@inproceedings{smartad2026,
  title = {{S}mart{AD}: Capacity-Aligned Agent Distillation for Small Language Models},
  author = {Tang, Guokai and Zhao, Feng},
  editor = {Liakata, Maria and Moreira, Viviane P. and Zhang, Jiajun and Jurgens, David},
  booktitle = {Findings of the {A}ssociation for {C}omputational {L}inguistics: {ACL} 2026},
  month = {jul},
  year = {2026},
  address = {San Diego, California, United States},
  publisher = {Association for Computational Linguistics},
  url = {https://aclanthology.org/2026.findings-acl.1349/},
  doi = {10.18653/v1/2026.findings-acl.1349},
  pages = {27045--27057},
  isbn = {979-8-89176-395-1}
}

@misc{dopd2026,
  title = {{DOPD}: Dual On-policy Distillation},
  author = {Yu, Xinlei and Li, Gen and Si, Qingyi and Zhang, Guibin and Xu, Yuqi and Wang, Congcong and Dong, Shuai and Tuo, Kaiwen and Zeng, Xiangyu and Feng, Kaituo and Wang, Qunzhong and Shi, Yang and Hu, Xiaobin and Yue, Xiangyu and Wang, Jiaqi and Yan, Shuicheng},
  year = {2026},
  eprint = {2606.30626},
  archiveprefix = {arXiv},
  url = {https://arxiv.org/abs/2606.30626}
}

@misc{tcod2026,
  title = {{TCOD}: Exploring Temporal Curriculum in On-Policy Distillation for Multi-turn Autonomous Agents},
  author = {Wang, Jiaqi and Zhang, Wenhao and Shi, Weijie and Li, Yaliang and Cheng, James},
  year = {2026},
  eprint = {2604.24005},
  archiveprefix = {arXiv},
  url = {https://arxiv.org/abs/2604.24005}
}

@misc{reopd2026,
  title = {Multi-Turn On-Policy Distillation with Prefix Replay},
  author = {Liao, Baohao and Dong, Hanze and Monz, Christof and Xu, Xinxing and Dong, Li and Wei, Furu},
  year = {2026},
  eprint = {2607.04763},
  archiveprefix = {arXiv},
  url = {https://arxiv.org/abs/2607.04763}
}

@misc{sageopd2026,
  title = {{SAGE-OPD}: Selective Agent-Guided Intervention for Multi-Turn On-Policy Distillation},
  author = {Zhou, Yuhang and Zhang, Lizhu and Wu, Yifan and Wang, Mingyi and Peng, Bo and Liu, Jiayi and Fan, Xiangjun and Zhao, Zhuokai},
  year = {2026},
  eprint = {2606.19659},
  archiveprefix = {arXiv},
  url = {https://arxiv.org/abs/2606.19659}
}

@misc{serl2026,
  title = {What and When to Distill: Selective Hindsight Distillation for Multi-Turn Agents},
  author = {Li, Xiaozhe and Lyu, Tianyi and Li, Yang and Ma, Yichuan and Li, Peiji and Li, Linyang and Guo, Qipeng and Lin, Dahua and Chen, Kai},
  year = {2026},
  eprint = {2605.19447},
  archiveprefix = {arXiv},
  url = {https://arxiv.org/abs/2605.19447}
}

@misc{stepopsd2026,
  title = {{StepOPSD}: Step-Aware Online Preference Self-Distillation for Agent Reinforcement Learning},
  author = {Zhang, Yanfei and Lin, Xu and Wu, Chenglin},
  year = {2026},
  eprint = {2605.27140},
  archiveprefix = {arXiv},
  url = {https://arxiv.org/abs/2605.27140}
}

@misc{hintsd2026,
  title = {{HINT-SD}: Targeted Hindsight Self-Distillation for Long-Horizon Agents},
  author = {Yeo, Woongyeong and Choi, Yumin and Ki, Taekyung and Hwang, Sung Ju},
  year = {2026},
  eprint = {2605.17873},
  archiveprefix = {arXiv},
  url = {https://arxiv.org/abs/2605.17873}
}

@misc{turnopd2026,
  title = {{TurnOPD}: Making On-Policy Distillation Turn-Aware for Efficient Long-Horizon Agent Training},
  author = {Zhou, Yuhang and Zheng, Kai and Li, Haoling and Peng, Dengyun and Xu, Can and Chen, Jingjing},
  year = {2026},
  eprint = {2607.05804},
  archiveprefix = {arXiv},
  url = {https://arxiv.org/abs/2607.05804}
}

@misc{grpo2024,
  title = {{DeepSeekMath}: Pushing the Limits of Mathematical Reasoning in Open Language Models},
  author = {Shao, Zhihong and Wang, Peiyi and Zhu, Qihao and Xu, Runxin and Song, Junxiao and Bi, Xiao and Zhang, Haowei and Zhang, Mingchuan and Li, Y. K. and Wu, Y. and Guo, Daya},
  year = {2024},
  eprint = {2402.03300},
  archiveprefix = {arXiv},
  url = {https://arxiv.org/abs/2402.03300}
}

@inproceedings{alfworld2021,
  title = {{ALFWorld}: Aligning Text and Embodied Environments for Interactive Learning},
  author = {Shridhar, Mohit and Yuan, Xingdi and C{\^o}t{\'e}, Marc-Alexandre and Bisk, Yonatan and Trischler, Adam and Hausknecht, Matthew},
  booktitle = {International Conference on Learning Representations},
  year = {2021},
  url = {https://openreview.net/forum?id=0IOX0YcCdTn}
}

@inproceedings{alfred2020,
  title = {{ALFRED}: A Benchmark for Interpreting Grounded Instructions for Everyday Tasks},
  author = {Shridhar, Mohit and Thomason, Jesse and Gordon, Daniel and Bisk, Yonatan and Han, Winson and Mottaghi, Roozbeh and Zettlemoyer, Luke and Fox, Dieter},
  booktitle = {Proceedings of the IEEE/CVF Conference on Computer Vision and Pattern Recognition},
  year = {2020},
  pages = {10740--10749},
  url = {https://openaccess.thecvf.com/content_CVPR_2020/html/Shridhar_ALFRED_A_Benchmark_for_Interpreting_Grounded_Instructions_for_Everyday_Tasks_CVPR_2020_paper.html}
}

@inproceedings{textworld2018,
  author = {C{\^o}t{\'e}, Marc-Alexandre and K{\'a}d{\'a}r, {\'A}kos and Yuan, Xingdi and Kybartas, Ben and Barnes, Tavian and Fine, Emery and Moore, James and Hausknecht, Matthew and El Asri, Layla and Adada, Mahmoud and Tay, Wendy and Trischler, Adam},
  editor = {Cazenave, Tristan and Saffidine, Abdallah and Sturtevant, Nathan},
  title = {{TextWorld}: A Learning Environment for Text-Based Games},
  booktitle = {Computer Games},
  year = {2019},
  publisher = {Springer International Publishing},
  address = {Cham},
  pages = {41--75},
  isbn = {978-3-030-24337-1},
  doi = {10.1007/978-3-030-24337-1_3},
  url = {https://link.springer.com/chapter/10.1007/978-3-030-24337-1_3},
  volume = {1017},
  series = {Communications in Computer and Information Science}
}

@inproceedings{webshop2022,
  author = {Yao, Shunyu and Chen, Howard and Yang, John and Narasimhan, Karthik},
  booktitle = {Advances in Neural Information Processing Systems},
  doi = {10.52202/068431-1508},
  editor = {S. Koyejo and S. Mohamed and A. Agarwal and D. Belgrave and K. Cho and A. Oh},
  pages = {20744--20757},
  publisher = {Curran Associates, Inc.},
  title = {{WebShop}: Towards Scalable Real-World Web Interaction with Grounded Language Agents},
  url = {https://proceedings.neurips.cc/paper_files/paper/2022/file/82ad13ec01f9fe44c01cb91814fd7b8c-Paper-Conference.pdf},
  volume = {35},
  year = {2022}
}

@inproceedings{react2023,
  title = {{ReAct}: Synergizing Reasoning and Acting in Language Models},
  author = {Yao, Shunyu and Zhao, Jeffrey and Yu, Dian and Du, Nan and Shafran, Izhak and Narasimhan, Karthik and Cao, Yuan},
  booktitle = {International Conference on Learning Representations},
  year = {2023},
  url = {https://openreview.net/forum?id=WE_vluYUL-X}
}

@inproceedings{reflexion2023,
  author = {Shinn, Noah and Cassano, Federico and Gopinath, Ashwin and Narasimhan, Karthik and Yao, Shunyu},
  booktitle = {Advances in Neural Information Processing Systems},
  doi = {10.52202/075280-0377},
  editor = {A. Oh and T. Naumann and A. Globerson and K. Saenko and M. Hardt and S. Levine},
  pages = {8634--8652},
  publisher = {Curran Associates, Inc.},
  title = {Reflexion: language agents with verbal reinforcement learning},
  url = {https://proceedings.neurips.cc/paper_files/paper/2023/file/1b44b878bb782e6954cd888628510e90-Paper-Conference.pdf},
  volume = {36},
  year = {2023}
}

@inproceedings{reflact2025,
  title = {{R}efl{A}ct: World-Grounded Decision Making in {LLM} Agents via Goal-State Reflection},
  author = {Kim, Jeonghye and Rhee, Sojeong and Kim, Minbeom and Kim, Dohyung and Lee, Sangmook and Sung, Youngchul and Jung, Kyomin},
  editor = {Christodoulopoulos, Christos and Chakraborty, Tanmoy and Rose, Carolyn and Peng, Violet},
  booktitle = {Proceedings of the 2025 Conference on Empirical Methods in Natural Language Processing},
  month = {nov},
  year = {2025},
  address = {Suzhou, China},
  publisher = {Association for Computational Linguistics},
  url = {https://aclanthology.org/2025.emnlp-main.1697/},
  doi = {10.18653/v1/2025.emnlp-main.1697},
  pages = {33433--33465},
  isbn = {979-8-89176-332-6}
}

@misc{fromhistorytostate2026,
  title = {From History to State: Constant-Context Skill Learning for {LLM} Agents},
  author = {Xie, Haoyang and Wang, Xinyuan and Wang, Yancheng and Zhao, Puda and Ju, Feng},
  year = {2026},
  eprint = {2605.05413},
  archiveprefix = {arXiv},
  url = {https://arxiv.org/abs/2605.05413}
}

@misc{qwen32025,
  title = {{Qwen3} Technical Report},
  author = {An Yang and Anfeng Li and Baosong Yang and Beichen Zhang and Binyuan Hui and Bo Zheng and Bowen Yu and Chang Gao and Chengen Huang and Chenxu Lv and Chujie Zheng and Dayiheng Liu and Fan Zhou and Fei Huang and Feng Hu and Hao Ge and Haoran Wei and Huan Lin and Jialong Tang and Jian Yang and Jianhong Tu and Jianwei Zhang and Jianxin Yang and Jiaxi Yang and Jing Zhou and Jingren Zhou and Junyang Lin and Kai Dang and Keqin Bao and Kexin Yang and Le Yu and Lianghao Deng and Mei Li and Mingfeng Xue and Mingze Li and Pei Zhang and Peng Wang and Qin Zhu and Rui Men and Ruize Gao and Shixuan Liu and Shuang Luo and Tianhao Li and Tianyi Tang and Wenbiao Yin and Xingzhang Ren and Xinyu Wang and Xinyu Zhang and Xuancheng Ren and Yang Fan and Yang Su and Yichang Zhang and Yinger Zhang and Yu Wan and Yuqiong Liu and Zekun Wang and Zeyu Cui and Zhenru Zhang and Zhipeng Zhou and Zihan Qiu},
  year = {2025},
  eprint = {2505.09388},
  archiveprefix = {arXiv},
  url = {https://arxiv.org/abs/2505.09388}
}

@misc{qwen252024,
  title = {{Qwen2.5} Technical Report},
  author = {An Yang and Baosong Yang and Beichen Zhang and Binyuan Hui and Bo Zheng and Bowen Yu and Chengyuan Li and Dayiheng Liu and Fei Huang and Haoran Wei and Huan Lin and Jian Yang and Jianhong Tu and Jianwei Zhang and Jianxin Yang and Jiaxi Yang and Jingren Zhou and Junyang Lin and Kai Dang and Keming Lu and Keqin Bao and Kexin Yang and Le Yu and Mei Li and Mingfeng Xue and Pei Zhang and Qin Zhu and Rui Men and Runji Lin and Tianhao Li and Tingyu Xia and Xingzhang Ren and Xuancheng Ren and Yang Fan and Yang Su and Yichang Zhang and Yu Wan and Yuqiong Liu and Zeyu Cui and Zhenru Zhang and Zihan Qiu},
  year = {2024},
  eprint = {2412.15115},
  archiveprefix = {arXiv},
  url = {https://arxiv.org/abs/2412.15115}
}

@inproceedings{expel2024,
  title = {{ExpeL}: {LLM} Agents Are Experiential Learners},
  author = {Zhao, Andrew and Huang, Daniel and Xu, Quentin and Lin, Matthieu and Liu, Yong-Jin and Huang, Gao},
  booktitle = {Proceedings of the AAAI Conference on Artificial Intelligence},
  year = {2024},
  pages = {19632--19642},
  doi = {10.1609/aaai.v38i17.29936},
  volume = {38},
  url = {https://ojs.aaai.org/index.php/AAAI/article/view/29936}
}

@inproceedings{awm2025,
  title = {Agent Workflow Memory},
  author = {Wang, Zora Zhiruo and Mao, Jiayuan and Fried, Daniel and Neubig, Graham},
  booktitle = {Proceedings of the 42nd International Conference on Machine Learning},
  year = {2025},
  volume = {267},
  series = {Proceedings of Machine Learning Research},
  pages = {63897--63911},
  publisher = {PMLR},
  url = {https://proceedings.mlr.press/v267/wang25bx.html}
}

@inproceedings{synapse2024,
  author = {Zheng, Longtao and Wang, Rundong and Wang, Xinrun and An, Bo},
  booktitle = {International Conference on Learning Representations},
  editor = {B. Kim and Y. Yue and S. Chaudhuri and K. Fragkiadaki and M. Khan and Y. Sun},
  pages = {19036--19066},
  title = {Synapse: Trajectory-as-Exemplar Prompting with Memory for Computer Control},
  url = {https://proceedings.iclr.cc/paper_files/paper/2024/file/52f050499cf82fa8efb588e263f6f3a7-Paper-Conference.pdf},
  volume = {2024},
  year = {2024}
}

@inproceedings{stateact2025,
  title = {{S}tate{A}ct: Enhancing {LLM} Base Agents via Self-prompting and State-tracking},
  author = {Rozanov, Nikolai and Rei, Marek},
  editor = {Kamalloo, Ehsan and Gontier, Nicolas and Lu, Xing Han and Dziri, Nouha and Murty, Shikhar and Lacoste, Alexandre},
  booktitle = {Proceedings of the 1st Workshop for Research on Agent Language Models (REALM 2025)},
  month = {jul},
  year = {2025},
  address = {Vienna, Austria},
  publisher = {Association for Computational Linguistics},
  url = {https://aclanthology.org/2025.realm-1.27/},
  doi = {10.18653/v1/2025.realm-1.27},
  pages = {367--385},
  isbn = {979-8-89176-264-0}
}

@article{zipact2026,
title={ZipAct: Zipping Interaction History into a Compact State for Efficient {LLM} Agents},
author={Zhiming Pan and Junyu Luo and Zhiping Xiao and Kaize Ding and Xiao Luo and Ming Zhang},
journal={Transactions on Machine Learning Research},
issn={2835-8856},
year={2026},
url={https://openreview.net/forum?id=ZssIalqqrz},
}

@inproceedings{samem2026,
  title = {{SAM}em: State-Aware Memory as a Fine-Grained Memory for {LLM} Agents in Decision-Making},
  author = {Wang, Tong and Xu, Pei and Cao, Shiyue and Yang, Likun and Li, Daipeng and Jiao, Jianbin and Huang, Kaiqi},
  editor = {Liakata, Maria and Moreira, Viviane P. and Zhang, Jiajun and Jurgens, David},
  booktitle = {Findings of the {A}ssociation for {C}omputational {L}inguistics: {ACL} 2026},
  month = {jul},
  year = {2026},
  address = {San Diego, California, United States},
  publisher = {Association for Computational Linguistics},
  url = {https://aclanthology.org/2026.findings-acl.722/},
  doi = {10.18653/v1/2026.findings-acl.722},
  pages = {14691--14710},
  isbn = {979-8-89176-395-1}
}

@inproceedings{proxystate2026,
  title = {Toward Scalable Verifiable Reward: Proxy State-Based Evaluation for Multi-turn Tool-Calling {LLM} Agents},
  author = {Chuang, Yun-Shiuan and Kulkarni, Chaitanya and Chiu, Alec M. and Thangali, Avinash and Pan, Zijie and Shekhar, Shivani and Ge, Yirou and Li, Yixi and Kona, Uma and Pang, Linsey and Mehrotra, Prakhar},
  editor = {Li, Yunyao and Rehm, Georg and Tu, Mei},
  booktitle = {Proceedings of the 64th Annual Meeting of the {A}ssociation for {C}omputational {L}inguistics (Volume 6: Industry Track)},
  month = {jul},
  year = {2026},
  address = {San Diego, California, USA},
  publisher = {Association for Computational Linguistics},
  url = {https://aclanthology.org/2026.acl-industry.87/},
  doi = {10.18653/v1/2026.acl-industry.87},
  pages = {1251--1264},
  isbn = {979-8-89176-394-4}
}

@misc{aropd2026,
  title = {Beyond Absolute Imitation: Anchored Residual Guidance for Privileged On-Policy Distillation},
  author = {Zhang, Wenhao},
  year = {2026},
  eprint = {2606.10385},
  archiveprefix = {arXiv},
  url = {https://arxiv.org/abs/2606.10385}
}

@misc{topd2026,
  title = {Bridging Reasoning Trajectories in On-Policy Distillation via Near-Future Guidance},
  author = {Jiang, Yuxuan and Ferraro, Francis},
  year = {2026},
  eprint = {2606.00305},
  archiveprefix = {arXiv},
  url = {https://arxiv.org/abs/2606.00305}
}

@inproceedings{ftrl2026,
  title = {Feedback-Driven Tool-Use Improvements in Large Language Models via Automated Build Environments},
  author = {Ye, Junjie and Jiang, Changhao and Du, Zhengyin and Xu, Yufei and Yao, Xuesong and Xi, Zhiheng and Fan, Xiaoran and Zhang, Qi and Gui, Tao and Huang, Xuanjing and Chen, Jiecao},
  editor = {Liakata, Maria and Moreira, Viviane P. and Zhang, Jiajun and Jurgens, David},
  booktitle = {Findings of the {A}ssociation for {C}omputational {L}inguistics: {ACL} 2026},
  month = {jul},
  year = {2026},
  address = {San Diego, California, United States},
  publisher = {Association for Computational Linguistics},
  url = {https://aclanthology.org/2026.findings-acl.109/},
  doi = {10.18653/v1/2026.findings-acl.109},
  pages = {2293--2323},
  isbn = {979-8-89176-395-1}
}

@inproceedings{toolhop2025,
  title = {{T}ool{H}op: A Query-Driven Benchmark for Evaluating Large Language Models in Multi-Hop Tool Use},
  author = {Ye, Junjie and Du, Zhengyin and Yao, Xuesong and Lin, Weijian and Xu, Yufei and Chen, Zehui and Wang, Zaiyuan and Zhu, Sining and Xi, Zhiheng and Yuan, Siyu and Gui, Tao and Zhang, Qi and Huang, Xuanjing and Chen, Jiecao},
  editor = {Che, Wanxiang and Nabende, Joyce and Shutova, Ekaterina and Pilehvar, Mohammad Taher},
  booktitle = {Proceedings of the 63rd Annual Meeting of the Association for Computational Linguistics (Volume 1: Long Papers)},
  month = {jul},
  year = {2025},
  address = {Vienna, Austria},
  publisher = {Association for Computational Linguistics},
  url = {https://aclanthology.org/2025.acl-long.150/},
  doi = {10.18653/v1/2025.acl-long.150},
  pages = {2995--3021},
  isbn = {979-8-89176-251-0}
}

@inproceedings{bfcl2025,
  title = {The Berkeley Function Calling Leaderboard ({BFCL}): From Tool Use to Agentic Evaluation of Large Language Models},
  author = {Patil, Shishir G and Mao, Huanzhi and Yan, Fanjia and Ji, Charlie Cheng-Jie and Suresh, Vishnu and Stoica, Ion and Gonzalez, Joseph E.},
  booktitle = {Proceedings of the 42nd International Conference on Machine Learning},
  pages = {48371--48392},
  year = {2025},
  editor = {Singh, Aarti and Fazel, Maryam and Hsu, Daniel and Lacoste-Julien, Simon and Berkenkamp, Felix and Maharaj, Tegan and Wagstaff, Kiri and Zhu, Jerry},
  volume = {267},
  series = {Proceedings of Machine Learning Research},
  month = {13--19 Jul},
  publisher = {PMLR},
  url = {https://proceedings.mlr.press/v267/patil25a.html}
}

@misc{dartsd2026,
  title = {{DART-SD}: Diamond-topology Aware Retrieval and Tuning for Self-Distillation of Multi-Turn Tool-Calling Agents},
  author = {Xu, Hangrui and Wang, Jiarui and Yang, Yang and Zhu, Chuanbo and Chen, Fangda and Wu, Ziqi and Cai, Jingming and Song, Yan},
  year = {2026},
  eprint = {2608.18524},
  archiveprefix = {arXiv},
  url = {https://arxiv.org/abs/2608.18524}
}

@misc{gcopd2026,
  title = {Graph-Conditioned On-Policy Agent Distillation from Off-the-Shelf Teachers},
  author = {Yi, Xiaohan and Luo, Wen and Huang, Yani and Zhan, Junfeng and Qin, Asher and Zhao, Peilin and Xiao, Xi},
  year = {2026},
  eprint = {2609.37522},
  archiveprefix = {arXiv},
  url = {https://arxiv.org/abs/2609.37522}
}

@misc{alignopsd2026,
  title = {Beyond Timestamps: Decision-Aligned On-Policy Distillation for Long-Horizon Agents},
  author = {Chen, Mingju and Lv, Can and Liu, Jinrong and Zhang, Huan and Chang, Heng and Zhou, Shiji},
  year = {2026},
  eprint = {2609.33391},
  archiveprefix = {arXiv},
  url = {https://arxiv.org/abs/2609.33391}
}

@misc{retireopd2026,
  title = {{RetireOPD}: Self-Retiring On-Policy Distillation for Agentic Reinforcement Learning},
  author = {Yu, Yan and Lu, Zhengxi and Liu, Yizhou and Pan, Yichen and Wang, Aozhe and Chen, Qipeng and Yang, Hua and Zhang, Wenqi and Chen, Qianglong and Shen, Yongliang},
  year = {2026},
  eprint = {2609.20784},
  archiveprefix = {arXiv},
  url = {https://arxiv.org/abs/2609.20784}
}

@misc{actfirst2026,
  title = {Act First, Reason Later: Accelerating On-Policy Distillation for Multi-Turn Agents via Reference-Conditioned Inverse Dynamics},
  author = {Zheng, Zubin and Wu, Jiahao and Zhang, Shaofeng and Zhang, Zhirui and Ong, Yew-Soon and Liu, Shengcai},
  year = {2026},
  eprint = {2609.36608},
  archiveprefix = {arXiv},
  url = {https://arxiv.org/abs/2609.36608}
}

@inproceedings{gigpo2025,
 author = {Feng, Lang and Xue, Zhenghai and Liu, Tingcong and An, Bo},
 booktitle = {Advances in Neural Information Processing Systems},
 doi = {10.52202/085713-1544},
 editor = {D. Belgrave and C. Zhang and H. Lin and R. Pascanu and P. Koniusz and M. Ghassemi and N. Chen},
 pages = {46375--46408},
 publisher = {Curran Associates, Inc.},
 title = {Group-in-Group Policy Optimization for {LLM} Agent Training},
 url = {https://proceedings.neurips.cc/paper_files/paper/2025/file/420c9f777c0b4f78d515e53cf74d58b2-Paper-Conference.pdf},
 volume = {38, Main Conference},
 year = {2025}
}

@inproceedings{hgpo2026,
 author = {He, Shuo and Feng, Lang and Wei, Qi and Cheng, Xin and Feng, Lei and An, Bo},
 booktitle = {International Conference on Learning Representations},
 editor = {C. Vondrick and B. Hariharan and C. Raffel and L. Pinto and D. Yang and A. Faust},
 pages = {27572--27593},
 title = {Hierarchy-of-Groups Policy Optimization for Long-Horizon Agentic Tasks},
 url = {https://proceedings.iclr.cc/paper_files/paper/2026/file/2e7eb932f93d5db3b61c7a001e72fd0f-Paper-Conference.pdf},
 volume = {2026},
 year = {2026}
}

@misc{rlsd2026,
  title = {Self-Distilled {RLVR}},
  author = {Yang, Chenxu and Qin, Chuanyu and Si, Qingyi and Chen, Minghui and Gu, Naibin and Yao, Dingyu and Lin, Zheng and Wang, Weiping and Wang, Jiaqi and Duan, Nan},
  year = {2026},
  eprint = {2604.03128},
  archiveprefix = {arXiv},
  url = {https://arxiv.org/abs/2604.03128}
}

@misc{sod2026,
  title = {{SOD}: Step-wise On-policy Distillation for Small Language Model Agents},
  author={Qiyong Zhong and Mao Zheng and Mingyang Song and Xin Lin and Jie Sun and Houcheng Jiang and Xiang Wang and Junfeng Fang},
  year={2026},
  eprint={2605.07725},
  archivePrefix={arXiv},
  primaryClass={cs.CL},
  url={https://arxiv.org/abs/2605.07725}, 
}

@inproceedings{toolrl2025,
  title = {{ToolRL}: Reward is All Tool Learning Needs},
  author = {Qian, Cheng and Acikgoz, Emre Can and He, Qi and Wang, Hongru and Chen, Xiusi and Hakkani-T{\"u}r, Dilek and Tur, Gokhan and Ji, Heng},
  booktitle = {Advances in Neural Information Processing Systems},
  year = {2025},
  url = {https://papers.nips.cc/paper_files/paper/2025/hash/97c5b2707228e7e3fb67e4ecc2e0e607-Abstract-Conference.html}
}

@inproceedings{matchtir2026,
  title = {{MatchTIR}: Fine-Grained Supervision for Tool-Integrated Reasoning via Bipartite Matching},
  author = {Qu, Changle and Dai, Sunhao and Cai, Hengyi and Xu, Jun and Wang, Shuaiqiang and Yin, Dawei},
  editor = {Liakata, Maria and Moreira, Viviane P. and Zhang, Jiajun and Jurgens, David},
  booktitle = {Proceedings of the 64th Annual Meeting of the Association for Computational Linguistics (Volume 1: Long Papers)},
  month = {jul},
  year = {2026},
  address = {San Diego, California, United States},
  publisher = {Association for Computational Linguistics},
  url = {https://aclanthology.org/2026.acl-long.549/},
  doi = {10.18653/v1/2026.acl-long.549},
  pages = {11953--11968},
  isbn = {979-8-89176-390-6}
}

\clearpage
\appendix
\section{Training and Evaluation}
\label{sec:hyperparameters}

\subsection{Shared Training Settings}
Following the base GRPO and self-distillation settings of prior work~\cite{sdar2026,ocsd2026}, we use a common setup across Qwen3-1.7B, Qwen2.5-3B-Instruct, and Qwen3-4B, with shared optimization settings in Table~\ref{tab:supp_common_hparams}. Every rollout starts from the task's initial state, and references are seen only by the teacher. \fullpath{} supplies the reference on all turns, while \method{} applies contextualized guidance only on matched turns. Qwen3-4B settings additionally follow the compared 4B baselines (Appendix~\ref{sec:4b_protocols}).

\begin{table}[!ht]
\centering
\begingroup
\fontsize{10}{12}\selectfont
\setlength{\tabcolsep}{1.5mm}
\begin{tabular}{@{}P{0.39\columnwidth}P{0.53\columnwidth}@{}}
\toprule
Parameter & Setting \\
\midrule
Learning rate & $10^{-6}$ \\
Rollouts per task & 8 \\
Tasks per update & 16 (ALFWorld/WebShop); 32 (FTRL) \\
Entropy coefficient & $0.001$ \\
SDL coefficient & $0.01$ \\
SDL estimator & Chosen-token K3, original response-token denominator \\
Reference KL & None \\
\bottomrule
\end{tabular}
\endgroup
\caption{Shared optimization settings for local runs.}
\label{tab:supp_common_hparams}
\end{table}

Table~\ref{tab:supp_domain_hparams} summarizes update budgets, episode horizons, and context limits. On ALFWorld, Qwen3-1.7B uses 250 updates and a 30-turn horizon; Qwen2.5-3B and Qwen3-4B use 150 updates and a 50-turn horizon. The smaller 1.7B model has not converged after 150 updates across methods (Figure~\ref{fig:training_curves}), so all its compared methods receive 250 updates to approach convergence. WebShop uses a common 150-update, 15-turn setting across models. FTRL uses three epochs, or 207 updates, with episode Solve-F1 as the GRPO reward. Trained models are evaluated at the last checkpoint of each run, without checkpoint selection.

\begin{table*}[t]
\centering
\begingroup
\fontsize{10}{12}\selectfont
\setlength{\tabcolsep}{2mm}
\begin{tabular*}{\textwidth}{@{\extracolsep{\fill}}llrrrr@{}}
\toprule
Environment & Model & Updates & Max turns & Prompt & Response \\
\midrule
ALFWorld & Qwen3-1.7B & 250 & 30 & 2,048 & 512 \\
ALFWorld & Qwen2.5-3B / Qwen3-4B & 150 & 50 & 2,048 & 1,024 / 512 \\
WebShop & All three models & 150 & 15 & 4,096 & 512 \\
FTRL & Qwen3-4B & 207 & 10 & 12,288 & 4,096 \\
\bottomrule
\end{tabular*}
\endgroup
\caption{Update budgets, episode horizons, and student prompt/response token limits.}
\label{tab:supp_domain_hparams}
\end{table*}

ALFWorld and WebShop use two-turn history, training temperature $1.0$, and evaluation temperature $0.4$. FTRL retains full history, allows 24,576 teacher-prompt tokens, and disables native thinking for training and in-domain evaluation while requesting brief reasoning before tool calls. For Qwen3-4B, Vanilla, \fullpath{}, and \method{} use seeds 0, 22, and 42 on ALFWorld and WebShop; other Qwen3-4B experiments use seed 42. Qwen3-1.7B and Qwen2.5-3B use seed 0. The code accompanying the submission will include complete configurations, prompts, launch scripts, and environment dependencies.

\paragraph{Hardware and training time.}
Qwen3-4B experiments use eight Kunlun P800 accelerators with 96\,GB memory each; Qwen3-1.7B and Qwen2.5-3B experiments use four NVIDIA H800 GPUs. We use PyTorch FSDP for actor and teacher workers and vLLM for rollouts. As a rough indication of training cost, representative ALFWorld runs take about 2--3 minutes per update for Qwen3-1.7B and 10--14 minutes for Qwen3-4B; Qwen3-4B WebShop runs take about 5 minutes per update. These wall-clock averages include periodic evaluation and vary with rollout length and environment.

\subsection{Evaluation and Baseline Sources}
\label{sec:4b_protocols}
We evaluate our trained ALFWorld and WebShop policies on 128 fixed held-out tasks with four full-start attempts each. Average is task-averaged success (Avg@4); WebShop additionally reports graded reward as Score. Avg@4 averages success over the four independently sampled attempts rather than taking the best of them, so it estimates the same single-attempt success rate as a one-attempt evaluation, with lower sampling variance. FTRL uses one greedy attempt on each of 200 held-out tasks at update 207, reporting Solve-P/R/F1 for required tool information. No evaluation uses references, state matching, or teacher guidance.

For transfer, we evaluate the final FTRL policies on BFCL and ToolHop without further training, following the evaluation settings of TurnSight~\cite{turnsight2026}.

\paragraph{Alignment with reported 4B baselines.}
Our Qwen3-4B experiments follow the published setups of the compared methods as closely as possible. On ALFWorld and WebShop, they match the public OCSD 4B recipe~\cite{ocsd2026} in update budget and horizon (150 updates; 50 turns on ALFWorld and 15 on WebShop), learning rate, gradient and PPO clipping ($1.0$ and $\pm0.2$), entropy coefficient, and invalid-action penalty ($0.1$). OCSD additionally uses a reference-KL term and its own reasoning-tag prompt, and evaluates one attempt per instance; our Avg@4 estimates the same single-attempt success rate from four attempts. On FTRL, we use the same 2,215/200 training and test split as TurnSight~\cite{turnsight2026}, its learning rate and 10-turn horizon, and 32 tasks per update for three epochs (207 updates).

\paragraph{Baseline sources.}
We evaluate all Vanilla baselines ourselves using the untrained base models. We also run every \fullpath{} and \method{} experiment and all Qwen3-1.7B methods. Our Qwen3-1.7B OCSD reproduction uses the shared training budgets and Avg@4 evaluation, with structured future-observation anchors on WebShop. The remaining baseline results are taken from the original papers: Qwen2.5-3B GRPO, Skill-SD, and SDAR from \citet{sdar2026}; Qwen3-4B ALFWorld/WebShop baselines from \citet{ocsd2026}, Table~1; and FTRL and transfer baselines from \citet{turnsight2026}, Table~1.

BFCL Avg is $0.5\,\mathrm{MT}+0.5\,\mathrm{Agentic}$, where MT averages the four multi-turn categories and Agentic averages Search and Memory. Overall averages FTRL mean(P,R,F1), BFCL Avg, and ToolHop accuracy. Local aggregates use unrounded values; published results retain their reported aggregates.

\subsection{Results Across Random Seeds}
\label{sec:multiseed_results}
Table~\ref{tab:multiseed_results} gives the complete Qwen3-4B results underlying Table~\ref{tab:main}. For Vanilla, \fullpath{} and \method{}, we report means and sample standard deviations across three seeds (0, 22, and 42). Results for GRPO through OCSD are reproduced from \citet{ocsd2026}, Table~1.

% Full-precision values and sources: ../table_results.json, qwen3_4b_multiseed.
\begin{table*}[t]
\centering
\begingroup
\fontsize{10}{12}\selectfont
\setlength{\tabcolsep}{2pt}
\resizebox{\textwidth}{!}{%
\begin{tabular}{@{}l*{9}{c}@{}}
\toprule
& \multicolumn{7}{c}{ALFWorld} & \multicolumn{2}{c}{WebShop} \\
\cmidrule(lr){2-8}\cmidrule(lr){9-10}
Method & Pick & Look & Clean & Heat & Cool & Pick2 & Average & Score & Acc \\
\midrule
\rowcolor{gray!15}
\multicolumn{10}{l}{\textbf{Qwen3-4B}} \\
Vanilla & $57.8{\pm}1.6$ & $32.6{\pm}1.2$ & $15.4{\pm}0.0$ & $20.6{\pm}7.5$ & $7.9{\pm}5.5$ & $20.1{\pm}6.8$ & $27.8{\pm}1.2$ & $14.7{\pm}0.4$ & $1.6{\pm}0.6$ \\
GRPO & $86.6{\pm}0.6$ & $59.5{\pm}5.1$ & $77.7{\pm}4.7$ & $67.1{\pm}4.1$ & $55.5{\pm}4.1$ & $55.0{\pm}6.2$ & $70.6{\pm}1.3$ & $81.9{\pm}3.6$ & $69.5{\pm}1.5$ \\
OPSD & $69.9{\pm}3.1$ & $39.2{\pm}4.5$ & $26.6{\pm}4.8$ & $26.6{\pm}4.3$ & $19.0{\pm}4.6$ & $26.4{\pm}3.8$ & $36.2{\pm}1.9$ & $27.8{\pm}2.9$ & $4.7{\pm}0.6$ \\
GRPO+OPSD & $88.0{\pm}0.4$ & $52.8{\pm}3.9$ & $75.0{\pm}4.1$ & $62.5{\pm}3.4$ & $68.3{\pm}8.1$ & $36.9{\pm}2.5$ & $68.8{\pm}2.6$ & $82.5{\pm}1.0$ & $71.3{\pm}1.2$ \\
RLSD & $87.9{\pm}4.0$ & $\mathbf{83.4}{\pm}1.8$ & $80.5{\pm}0.9$ & $70.3{\pm}9.4$ & $79.6{\pm}4.0$ & $60.6{\pm}4.3$ & $79.2{\pm}1.0$ & $83.7{\pm}1.5$ & $68.5{\pm}1.2$ \\
SDAR & $85.2{\pm}4.8$ & $63.9{\pm}4.0$ & $76.8{\pm}5.8$ & $80.0{\pm}3.6$ & $72.9{\pm}4.4$ & $42.6{\pm}6.9$ & $73.2{\pm}1.3$ & $83.6{\pm}1.8$ & $71.9{\pm}0.7$ \\
OCSD & $\mathbf{91.9}{\pm}1.4$ & $\underline{77.2}{\pm}4.4$ & $\underline{90.2}{\pm}4.6$ & $84.7{\pm}4.4$ & $\underline{80.9}{\pm}4.3$ & $\underline{63.8}{\pm}0.9$ & $\underline{82.8}{\pm}0.7$ & $\mathbf{86.9}{\pm}1.7$ & $\underline{73.7}{\pm}1.6$ \\
\fullpath{} & $89.3{\pm}0.5$ & $76.4{\pm}9.8$ & $\mathbf{90.4}{\pm}7.6$ & $\underline{87.8}{\pm}5.4$ & $79.4{\pm}14.1$ & $52.7{\pm}8.8$ & $80.2{\pm}2.0$ & $82.4{\pm}5.1$ & $73.6{\pm}2.9$ \\
\midrule
\method{} (ours) & $\underline{91.4}{\pm}3.6$ & $76.4{\pm}10.3$ & $88.5{\pm}7.6$ & $\mathbf{90.0}{\pm}2.9$ & $\mathbf{85.3}{\pm}5.6$ & $\mathbf{65.5}{\pm}11.8$ & $\mathbf{83.8}{\pm}3.3$ & $\underline{86.0}{\pm}0.7$ & $\mathbf{74.3}{\pm}0.5$ \\
\bottomrule
\end{tabular}%
}
\endgroup
\caption{Complete Qwen3-4B results (\%): mean $\pm$ standard deviation across three seeds. GRPO through OCSD are from \citet{ocsd2026}; Vanilla summarizes base-model evaluations. Bold/underline: best/second-best mean.}
\label{tab:multiseed_results}
\end{table*}

\section{References and State Interfaces}

\subsection{Reference Construction}
\label{sec:reference_construction}

Each training task retrieves one canonical reference by task identity. Only the synchronized teacher receives privileged context after rollout; the student always acts from the ordinary prompt. Evaluation retrieves no references. Dataset splits and evaluation settings are summarized in Appendix~\ref{sec:4b_protocols}.

\paragraph{ALFWorld references.}
We use the benchmark-provided TextWorld expert walkthrough associated with each training game and verify its execution before indexing it \cite{alfworld2021,textworld2018}. The index contains one reference for each of 3,553 training games. References average $6.00$ actions (median $6$, range $3$--$10$). For example, a heat-and-place task can yield
\begin{quote}
\texttt{go to cabinet 1 $\rightarrow$ open cabinet 1 $\rightarrow$}\\
\texttt{take mug 1 $\rightarrow$ go to microwave 1 $\rightarrow$}\\
\texttt{heat mug 1 $\rightarrow$ go to shelf 1 $\rightarrow$}\\
\texttt{put mug 1 in/on shelf 1}.
\end{quote}
The exact object and receptacle identifiers are task-specific.

\paragraph{WebShop references.}
We use the 1,000-product WebShop-small setting~\cite{webshop2022}, following the public environment configurations of SDAR and OCSD~\cite{sdar2026,ocsd2026}. For each of 6,910 goals in its fixed manifest, we deterministically construct a product-and-option trace from the goal's target ASIN and requested options:
\begin{quote}
\texttt{search[full product name] $\rightarrow$ click[target ASIN]}\\
\texttt{$\rightarrow$ click[option value] $\rightarrow \cdots \rightarrow$ click[buy now]}.
\end{quote}
These traces average $5.08$ actions (range $3$--$6$). Training uses 6,410 goals; the first 500 goals are held out, with a fixed 128-task validation subset. Only training references enter the matcher and teacher.

\paragraph{FTRL references.}
We execute the dataset's reference tool calls to obtain their returned information and cumulative prefix states. The state representation binds tool names and arguments to canonicalized returns; it uses neither gold answers nor unobserved future information to construct the student state. Of 2,215 training tasks, 2,181 have usable references. The remaining 34 receive GRPO updates but no SDL for both local methods.

\subsection{State Matching and Guidance}
\label{sec:state_adapters}

The reference constructor uses action prefixes and observable outcomes; for tool use, these outcomes are obtained by executing the reference calls. The rendered summary describes the student's actual state and may omit matcher-only fields. \fullpath{} and \method{} share the closing instruction; \method{} adds the state summary and grounded candidate and restricts SDL to matched turns.

\begin{table*}[t]
\centering
\setlength{\tabcolsep}{4mm}
\begin{tabular}{P{0.45\textwidth}P{0.45\textwidth}}
\toprule
\fullpath{} & \method{} \\
\midrule
\ttfamily
[Privileged Path Information]\newline
Complete successful path for this task:\newline
\{ground-truth path\}\newline
The current state may be on or off this path.\newline
Use the path as privileged guidance, but reason from the current observation and admissible actions.\newline
[/Privileged Path Information] &
\ttfamily
[Privileged Path Information]\newline
Complete successful path for this task:\newline
\{ground-truth path\}\newline
Current state summary:\newline
\{current state summary\}\newline
Candidate next action for the current state:\newline
\{candidate action\}\newline
The current state may be on or off this path.\newline
Use the path as privileged guidance, but reason from the current observation and admissible actions.\newline
[/Privileged Path Information] \\
\bottomrule
\end{tabular}
\caption{Teacher-only prompts. The closing instruction is identical; \method{} changes the matcher-conditioned route and adds the two local fields.}
\label{tab:supp_prompts}
\end{table*}

\paragraph{Common interface.}
Each environment supplies the signature constructors $\phi^{\mathrm{stu}}$ and $\phi^{\mathrm{ref}}$, the compatibility relation $\models$, action grounding $\Gamma$, and $\operatorname{Render}$ (Section~\ref{sec:matching}). Only the state fields and rules are environment-specific.

\paragraph{ALFWorld state adapter.}
The adapter reconstructs $\sigma_t$ from ordinary observations and successful executed transitions, and derives every $\bar\sigma_k$ symbolically from the canonical action prefix, without hidden simulator state or a reference replay cache. Its fields are location, inventory, task-object locations, and task-object properties. Location is the destination of the latest successful \texttt{go to}; observation-based inference is used only before any successful navigation. \texttt{take}, \texttt{move}/\texttt{put}, and \texttt{drop} update inventory and object location; \texttt{clean}, \texttt{cool}, \texttt{heat}, and \texttt{slice} update task-object properties. Heating removes a prior \texttt{cool} fact and cooling removes a prior \texttt{heat} fact.

Reported ALFWorld runs use exact object identifiers. Location and inventory must equal the reference pre-action state. Each reference-required object location must agree with the current location of that object, while reference-required properties must be a subset of current progress. This directional relation permits additional compatible completed facts without omitting or contradicting a requirement of the proposed continuation. The reference next action must also ground to an action in the current admissible-action list. Among compatible pre-action states, the matcher selects the greatest reference position.

\paragraph{History matcher used in the comparison.}
The history matcher first requires exact normalized equality between the current observation and a reference observation and requires the reference next action to be currently admissible. It then ranks observation-compatible positions lexicographically by exact reference-history suffix match, reference-history subsequence match, common action-history prefix length, and finally later reference position. Either a suffix or subsequence match routes SDL with the same binary mask $w_t=1$; an initial-state observation hit or a history mismatch abstains ($w_t=0$). There is no confidence weighting in this comparison or in the reported runs. Inventory agreement is logged for audit but is not a selection condition for this matcher.

\paragraph{WebShop state adapter.}
The stable task identity contains the instruction, target ASIN, and requested options. The execution signature contains page type, current ASIN, and the set of selected options; search wording and result-page number are ignored. At the search home, an available search action and a concrete reference query yield a matched search-query summary without forcing unique wording. On a results page, matching requires the target ASIN to be currently clickable. On the target item page, selected options must be a compatible subset of the requested options. Remaining options are order-invariant: multiple remaining options produce the set of currently admissible option clicks, one remaining option produces a unique click, and no remaining option produces \texttt{click[buy now]} when it is admissible. A wrong product page, incompatible selected option, invisible target product, or unavailable required click is unmatched.

\paragraph{FTRL state adapter.}
The state is a deduplicated set of facts observed through tool calls. Each fact binds the tool name, its arguments, and its canonicalized return value. We fill omitted arguments with literal defaults available in the tool definition and deterministically normalize returned values. This preserves queried entities and result-list order while making the state independent of the order of equivalent calls.

We construct the student's state from its observed tool results and reference states from the cumulative returns of the executed reference prefixes. Matching requires equality of these sets. Thus an empty reference prefix matches only an empty student state. Candidates whose information has already been acquired are skipped; the most informative compatible prefix supplies the next candidate, with the latest position breaking ties. The teacher receives the full reference, the student's observed state, and the compatible candidate, following the same routing, guidance construction, and learning objective as the other environments.

\paragraph{Teacher-context rendering.}
Table~\ref{tab:supp_prompts} shows that \fullpath{} and \method{} share the same complete path and closing instruction. \method{} adds only the current-state summary and grounded local guidance on matched turns. The table shows the unique-candidate form used by ALFWorld and by unique WebShop states. More generally, the grounded candidate $\Gamma(\bar a_{k_t};\mathcal{A}_t)$, abbreviated $\bar a_{k_t}$ in the main text, is a nonempty set of compatible actions: a singleton for ALFWorld and unique WebShop states, and a larger set for order-invariant WebShop option states. For the latter, the candidate field serializes all currently compatible option clicks rather than designating one of them. At the search home, the reference query appears in the state summary because search wording is not unique.

\paragraph{WebShop serialized example.}
The WebShop renderer follows the same full-path, current-state, local-guidance, and closing-instruction scaffold as shown in Table~\ref{tab:supp_webshop_prompt}. A matched turn with one required option remaining is serialized as follows.

\begin{table}[t]
\centering
\setlength{\tabcolsep}{1.5mm}
\begin{tabular}{P{0.92\columnwidth}}
\toprule
\ttfamily
[Privileged Path Information]\newline
Complete successful path for this task:\newline
search[VanciLin Mens Casual Leather Fashion Slip-on Loafers]\newline
$\rightarrow$ click[B07S7HDC88] $\rightarrow$ click[blue137] $\rightarrow$ click[9] $\rightarrow$ click[buy now]\newline
Current state summary:\newline
product=B07S7HDC88;\newline
selected=color=blue137;\newline
remaining(any order)=size=9.\newline
Candidate next action for the current state:\newline
click[9]\newline
The current state may be on or off this path.\newline
Use the path as privileged guidance, but reason from the current observation and admissible actions.\newline
[/Privileged Path Information] \\
\bottomrule
\end{tabular}
\caption{Serialized WebShop matched-turn context for a one-option-remaining state. Its structure matches the ALFWorld \method{} context.}
\label{tab:supp_webshop_prompt}
\end{table}

The \method{} prompt does not separately list the matched reference prefix, executed prefix, or match declaration. Unmatched turns have $w_t=0$ and receive no SDL.

\paragraph{Latest-position tie-break.}
When more than one reference pre-action state is compatible, the matcher chooses the greatest path index. This selects the most advanced verified phase, avoids repeating completed subgoals, and reduces stale continuations. The rule does not skip an incompatible intermediate phase: every selected position must independently satisfy the state requirements and admissible-action check.

\section{Fixed-State Teacher Probes}
\label{sec:teacher_probe_details}

\paragraph{Probe conditions and metrics.}
The probes use the three conditions of the main text. \emph{Matched-agreeing} turns are matched states whose sampled action agrees with the reference continuation; \emph{mismatched} turns are unmatched states with a feasible sampled action from a successful trajectory; \emph{matched-divergent} turns are matched states whose sampled action differs from the reference continuation. The first two conditions give the action-score change $\Delta G_t$ (Equation~\ref{eq:sampled_action_shift}) in the top block of Table~\ref{tab:teacher_audit}, which measures support for the feasible sampled action. The third gives the preference-margin change $\Delta M_t$ (Equations~\ref{eq:teacher_margin}--\ref{eq:teacher_margin_shift}) in the bottom block, which measures preference for the compatible reference action over the sampled action. Table~\ref{tab:supp_teacher_ci} gives the corresponding confidence intervals.

\begin{table*}[!t]
\centering
\setlength{\tabcolsep}{1mm}
\begingroup
\fontsize{10}{12}\selectfont
\begin{tabular*}{\textwidth}{@{\extracolsep{\fill}}lrr@{}}
\toprule
Condition / teacher context & Mean & 95\% CI \\
\midrule
\multicolumn{3}{l}{\textit{Support for feasible actions: successful trajectories ($\Delta G_t$)}} \\
Matched-agreeing & $+0.017$ & $[+0.002,+0.038]$ \\
Mismatched & $-0.052$ & $[-0.134,+0.001]$ \\
\midrule
\multicolumn{3}{l}{\textit{Preference for reference actions: matched-divergent turns ($\Delta M_t$)}} \\
No privilege & $0.000$ & -- \\
Abstract skill & $+0.153$ & $[+0.100,+0.218]$ \\
Random successful path (same task family) & $+0.442$ & $[+0.322,+0.584]$ \\
Reference path & $+1.158$ & $[+0.938,+1.401]$ \\
Contextualized reference path & $\mathbf{+1.424}$ & $\mathbf{[+1.173,+1.694]}$ \\
\bottomrule
\end{tabular*}
\endgroup
\caption{Mean probe effects and 95\% confidence intervals for Table~\ref{tab:teacher_audit}. Each group has 400 anchors; intervals use a game-cluster bootstrap after averaging the two frozen scorers within each anchor.}
\label{tab:supp_teacher_ci}
\end{table*}

\paragraph{Fixed real-anchor suite.}
We construct the intervention manifest from GRPO full-start rollouts at updates 5, 100, 150, 200, and 250, using the same structured-state matcher as \method{}. It contains 1,200 turns from 458 trajectories and 79 training games. Here the candidate is the matcher-selected grounded next reference action. The suite contains 400 matched-agreeing turns, 400 matched-divergent turns balanced between successful and failed trajectories, and 400 mismatched turns from successful trajectories. Table~\ref{tab:supp_anchor_composition} gives the source-update composition; the main paper pools these updates rather than treating them as scorer checkpoints.

Every retained turn satisfies the following quality controls: the parsed action equals the executed action, is present in the prompt's admissible-action set, executes successfully in the environment, contains no trailing text after the action span, and produces observed progress. Matched-agreeing and mismatched turns additionally come from successful trajectories of at most ten executed actions. Terminal outcome is used only for sampling, not as an action-level correctness label.

\begin{table*}[t]
\centering
\setlength{\tabcolsep}{2.2mm}
\begin{tabular}{llrrrrrr}
\toprule
Condition & Trajectory & Update 5 & 100 & 150 & 200 & 250 & Total \\
\midrule
Matched-agreeing & Success & 43 & 61 & 93 & 72 & 131 & 400 \\
Matched-divergent & Success & 40 & 38 & 47 & 35 & 40 & 200 \\
Matched-divergent & Failure & 66 & 54 & 26 & 32 & 22 & 200 \\
Mismatched & Success & 15 & 78 & 92 & 94 & 121 & 400 \\
\midrule
Total & & 164 & 231 & 258 & 233 & 314 & 1,200 \\
\bottomrule
\end{tabular}
\caption{Composition of the fixed real-anchor intervention suite. ``Update'' identifies the rollout source, not the scorer used for teacher-context intervention.}
\label{tab:supp_anchor_composition}
\end{table*}

\begin{table*}[t]
\centering
\setlength{\tabcolsep}{0.7mm}
\begingroup
\fontsize{10}{12}\selectfont
\begin{tabular}{lrrrr}
\toprule
& \multicolumn{2}{c}{Pre-RL proxy} & \multicolumn{2}{c}{GRPO update 250} \\
\cmidrule(lr){2-3}\cmidrule(lr){4-5}
Teacher context & Failure & Success & Failure & Success \\
\midrule
No privilege & $-3.077$ / 0.0\% & $-2.969$ / 0.0\% & $-2.976$ / 5.0\% & $-2.873$ / 0.5\% \\
Abstract skill & $-2.922$ / 0.5\% & $-2.710$ / 1.0\% & $-2.886$ / 5.5\% & $-2.765$ / 2.0\% \\
Random successful path (same family) & $-2.669$ / 3.5\% & $-2.545$ / 3.0\% & $-2.460$ / 10.5\% & $-2.452$ / 3.5\% \\
Shuffled exact-task path & $-2.039$ / 11.0\% & $-2.097$ / 9.0\% & $-1.755$ / 20.0\% & $-1.998$ / 8.5\% \\
Reference path & $-1.937$ / 12.5\% & $-1.914$ / 10.5\% & $-1.616$ / 20.0\% & $-1.794$ / 12.5\% \\
Contextualized reference path & $\mathbf{-1.644}$ / \textbf{14.5\%} & $\mathbf{-1.691}$ / \textbf{12.0\%} & $\mathbf{-1.317}$ / \textbf{23.5\%} & $\mathbf{-1.545}$ / \textbf{16.0\%} \\
\bottomrule
\end{tabular}
\endgroup
\caption{Candidate-minus-sampled action margin / positive-margin rate on matched-divergent turns, split by trajectory outcome (200 each). Higher is better.}
\label{tab:supp_teacher_b}
\end{table*}

\begin{table*}
\centering
\setlength{\tabcolsep}{2mm}
\begin{tabular}{lrrrr}
\toprule
& \multicolumn{2}{c}{Matched-agreeing} & Mismatched & Difference \\
\cmidrule(lr){2-3}\cmidrule(lr){4-4}\cmidrule(lr){5-5}
Scorer & FullPath & \method{} & FullPath & FullPath \\
\midrule
Pre-RL proxy & $+0.01895$ & $+0.02196$ & $-0.05244$ & $+0.07140$ \\
GRPO update 250 & $+0.01558$ & $+0.01558$ & $-0.05236$ & $+0.06794$ \\
\bottomrule
\end{tabular}
\caption{Observed-action score changes $\Delta G$ relative to no privilege, with 400 turns per condition. A positive matched-minus-mismatched difference means the identical FullPath context is more favorable on matched turns.}
\label{tab:supp_teacher_ac}
\end{table*}

\paragraph{Scoring procedure and contexts.}
The intervention holds the reached state, ordinary prompt, and sampled response fixed and changes only training-time teacher context. We score the mean log probability over the parsed action span, including its delimiter tags. For matched-divergent turns, the diagnostic response retains the sampled response's reasoning, formatting, and all other tokens; only its action span is exactly replaced by the matched candidate. We then measure the candidate-minus-sampled margin $M_t(z)$ and its context-induced change $\Delta M_t(z)$ as defined in the main text. For matched-agreeing and mismatched turns, we score the change $\Delta G_t(z)$ in the observed action relative to no privilege. No response is regenerated and no policy update occurs.

Contexts include no privilege, abstract task skill, a random successful path from another task in the same family, a shuffled exact-task path, the exact-task reference path, and its contextualized form. The last two are the contexts used by \fullpath{} and \method{}, respectively. Matched-agreeing turns use no privilege and the two reference contexts; mismatched turns use no privilege and FullPath because no compatible candidate exists. We score all anchors with pretrained Qwen3-1.7B and the GRPO update-250 actor, average their effects within each anchor, and use 20,000 game-cluster bootstrap samples for the 95\% percentile intervals. Whole games are resampled together to retain dependence among their turns.

\paragraph{Matched-divergent turns.}
Table~\ref{tab:supp_teacher_b} reports the outcome split at both scorer endpoints. A cell is the raw candidate-minus-sampled action margin followed by the percentage of anchors on which that margin is positive. The raw margin remains negative in aggregate, partly because the fixed reasoning was produced for the sampled action. The relevant paired intervention is the change from the no-privilege row.

Pooling both outcomes, FullPath changes the margin by $+1.158$ relative to no privilege, and \method{} changes it by $+1.424$. The paired \method{}--FullPath difference is $+0.266$ with a game-cluster-bootstrap 95\% interval of $[+0.206,+0.328]$. The ordering $\method{}>\text{FullPath}>\text{shuffled path}>\text{other same-family path}$ holds for both outcomes at both scorer endpoints. Thus the corrective shift is not explained by generic skill text, task-family vocabulary, or merely listing the exact task's actions.

\paragraph{Matched versus mismatched turns.}
Table~\ref{tab:supp_teacher_ac} applies the same FullPath context to matched-agreeing and mismatched turns. Each condition is paired against its own no-privilege scores; the two conditions contain different turns.

The matched-minus-mismatched difference is positive at both scorer endpoints. After averaging each turn over the two scorers, it is $+0.070$ with a game-cluster-bootstrap 95\% interval of $[+0.012,+0.153]$. The matched-agreeing mean is small because many reference-agreeing actions already have near-zero action NLL, and adverse effects there are rare: under \method{}, only one of 800 turn--scorer evaluations decreases the agreeing action score by more than $0.01$ (0.1\%).

On mismatched turns, suppression is concentrated in a subset of turns. FullPath lowers the sampled-action score by more than $0.01$ on 8.3\% of them under the pre-RL proxy and 6.5\% under GRPO-250, and by more than $0.1$ on 6.0\% and 5.8\%, respectively, while the median change is near zero. The same reference therefore behaves differently at matched and mismatched states. Divergence from the reference does not imply an incorrect action: some successful matched-divergent turns use a different valid receptacle, and a mismatch can arise when a successful rollout uses a different object instance from the reference.

\paragraph{Audited examples.}
Table~\ref{tab:supp_probe_examples} illustrates the state and action conditions in the probe using manually inspected anchors. The alarmclock example shows how a successful student trajectory can fall outside its task's reference path. The tomato example shows a matched state where the reference supplies a direct continuation toward an unfinished subgoal, although the student takes a detour. Thus, the two probes test support for feasible actions and preference for a compatible continuation, respectively. The measured context effects for these conditions appear in Table~\ref{tab:supp_teacher_ci}.

\begin{table*}[t]
\centering
\begingroup
\fontsize{10}{12}\selectfont
\setlength{\tabcolsep}{2mm}
\begin{tabular}{@{}P{0.16\textwidth}P{0.38\textwidth}P{0.40\textwidth}@{}}
\toprule
Probe condition & Reached state and student action & Reference relation and diagnostic purpose \\
\midrule
Matched-agreeing & At the initial state, the student chooses \texttt{go to coffeetable 1}. & The matched candidate is the same action. This tests whether guidance preserves support for an already reference-consistent decision. \\
\midrule
Matched-divergent & The student holds an unclean tomato at \texttt{countertop 3} and chooses \texttt{go to microwave 1}; the trajectory eventually succeeds. & The compatible reference candidate is \texttt{go to sinkbasin 1}. This tests whether state context strengthens preference for the continuation that addresses the pending cleaning subgoal. \\
\midrule
Mismatched & A successful student trajectory moves \texttt{alarmclock 1}. & The reference instead uses \texttt{alarmclock 2}, leaving no compatible continuation for the reached object state. This tests whether that reference suppresses support for the student's feasible action. \\
\bottomrule
\end{tabular}
\endgroup
\caption{Real ALFWorld probe examples from the manual context audit. The rows illustrate the sampling conditions; aggregate score changes are reported in Table~\ref{tab:supp_teacher_ci}.}
\label{tab:supp_probe_examples}
\end{table*}

\section{Matcher and Routing Audits}
\label{sec:matcher_audit}

\paragraph{Routing controls.}
Following the main paper, \fullpath{} applies plain full-path context on all turns. Matched-turn routing uses the structured matcher and applies the same plain FullPath context only on matched turns. The same-count random control samples, at every update, as many turns as the matcher selects. Unmatched-turn routing applies SDL to all complementary unmatched turns. Dynamic Context uses SMRC context on matched turns and plain FullPath on unmatched turns; \method{} combines SMRC context with matched-only routing.

\paragraph{Distillation strength in the same-count control.}
All routing controls share $\lambda_{\mathrm{SDL}}$, the teacher context, and the SDL normalization, whose denominator is the original response-token count including routed-out turns. The total SDL strength of an update is therefore set by the response tokens it selects. At every update, the random selector samples the current batch's matched-turn count without replacement, which equalizes the number of distilled turns per update. The two runs then generate different trajectories, so the counts cannot coincide across whole runs, and their aggregate ratios differ slightly. Table~\ref{tab:supp_selection_mass} reports both turn and token ratios. The random control selects more response tokens than matched routing ($16.4\%$ versus $14.2\%$ on average; $31.3\%$ versus $27.3\%$ at the last update), so it receives at least the same distillation strength and still trails matched routing by $11.3$ Avg@4 points.

\begin{table}[t]
\centering
\setlength{\tabcolsep}{1mm}
\begingroup
\fontsize{10}{12}\selectfont
\begin{tabular}{@{}P{0.43\columnwidth}cc@{}}
\toprule
Route & Turns & Tokens \\
\midrule
Matched & $15.3\;(31.8)$ & $14.2\;(27.3)$ \\
Random, same count & $16.5\;(32.1)$ & $16.4\;(31.3)$ \\
\bottomrule
\end{tabular}
\endgroup
\caption{Training-update mean (final) selected-turn and selected-response-token ratios (\%) for the FullPath routing controls. The token statistic counts eligible sampled-response tokens.}
\label{tab:supp_selection_mass}
\end{table}

\subsection{ALFWorld Matcher and Replay}
\paragraph{Same-row selector comparison.}
We apply the history-based and structured-state matchers to the same 35,712 archived turns from 14 ALFWorld rollout dumps spanning several checkpoints and policies. Applying both selectors to every turn makes their coverage and action choices directly comparable.

\begin{table}[t]
\centering
\setlength{\tabcolsep}{1mm}
\begin{tabular}{@{}lrr@{}}
\toprule
Metric & History & Structured state \\
\midrule
Candidate rows & 5,496 & 7,203 \\
Coverage & 15.4\% & 20.2\% \\
Exclusive rows & 68 & 1,775 \\
\midrule
Overlap & \multicolumn{2}{c}{5,428} \\
Same action on overlap & \multicolumn{2}{c}{5,395 (99.4\%)} \\
\bottomrule
\end{tabular}
\caption{Identical-row comparison of history-based and structured-state matching.}
\label{tab:supp_matcher_coverage}
\end{table}

The 68 history-only rows have fully attributable structured-state rejection reasons: 63 inventory mismatches, four object-location mismatches, and one agent-location mismatch. Thus, most rejected history matches are not parser failures; they are cases where similar observations or action histories conceal a different execution state. Conversely, the 1,775 state-only rows arise when structured progress supports a continuation despite the absence of a sufficiently similar history suffix.

\paragraph{End-to-end matcher ablation.}
We next hold the SMRC teacher context and matched-only routing semantics fixed and change only the matcher. Both rows use the last (update-250) checkpoint and the same 512-rollout evaluation protocol.

\begin{table}[t]
\centering
\setlength{\tabcolsep}{1mm}
\begingroup
\fontsize{10}{12}\selectfont
\begin{tabular}{lrr}
\toprule
Matcher & Avg@4 & Pass@4 \\
\midrule
History & $75.6$ & $85.9$ \\
Structured state & $\mathbf{86.5}$ & $\mathbf{91.4}$ \\
\bottomrule
\end{tabular}
\endgroup
\caption{Online matcher ablation with the teacher context fixed (\%).}
\label{tab:supp_matcher_online}
\end{table}

\begin{table}[t]
\centering
\setlength{\tabcolsep}{1.2mm}
\begingroup
\fontsize{10}{12}\selectfont
\begin{tabular}{@{}lrrrr@{}}
\toprule
& \multicolumn{2}{c}{History} & \multicolumn{2}{c}{Structured state} \\
\cmidrule(lr){2-3}\cmidrule(lr){4-5}
Task family & Replayed & Success & Replayed & Success \\
\midrule
Look & 100 & 98 & 81 & 81 \\
Pick & 100 & 100 & 100 & 100 \\
Clean & 100 & 100 & 100 & 100 \\
Cool & 100 & 100 & 100 & 100 \\
Heat & 100 & 100 & 100 & 100 \\
Pick Two & 300 & 294 & 300 & 300 \\
\midrule
All & 800 & 792 & 781 & 781 \\
\bottomrule
\end{tabular}
\endgroup
\caption{Candidate-plus-suffix replay by matcher and ALFWorld task family. The history and structured-state columns use independently stratified matcher-specific samples.}
\label{tab:supp_exact_replay}
\end{table}

The history run selects approximately $8.3\%$ of turns on average ($7.7\%$ at the final update). Its lower final result shows that structured execution progress is useful beyond choosing nearly identical candidates on the overlap.

\begin{table*}[t]
\centering
\setlength{\tabcolsep}{2mm}
\begin{tabular}{P{0.21\textwidth}P{0.24\textwidth}P{0.24\textwidth}P{0.20\textwidth}}
\toprule
Reached execution state & History/reference suggestion & Structured-state outcome & Compatibility interpretation \\
\midrule
\texttt{lettuce 3} has already been placed in the fridge & Open \texttt{fridge 1} again & Go to \texttt{countertop 2} & Advance to the second-object phase instead of repeating the completed placement phase. \\
\texttt{fork 1} is already clean & Clean \texttt{fork 1} again & Navigate toward the target receptacle & The object-property signature prevents a repeated transformation. \\
\texttt{mug 4} is already hot & Heat the mug again & Go toward the target shelf & The \texttt{hot} property aligns the turn with the post-heating reference position. \\
The student is currently at \texttt{fridge 1} & Return the canonical initial action, \texttt{go to diningtable 2} & Abstain & Agent-location mismatch prevents an initial-state continuation from supervising the reached state. \\
\bottomrule
\end{tabular}
\caption{Representative history aliases and structured-state decisions from the ALFWorld audit.}
\label{tab:supp_matcher_examples}
\end{table*}

\paragraph{Representative audited cases.}
Table~\ref{tab:supp_matcher_examples} summarizes cases from the disagreement and rejection audit. The first three illustrate observation aliases: the visible scene or recent history remains compatible with an earlier phase even though a placement or transformation subgoal is already complete. The final row illustrates abstention when history similarity conflicts with the reached location.

\paragraph{Executable-continuation replay.}
Admissibility of the first candidate alone does not show that it preserves task progress. For the history matcher, we recompute matches on archived rollouts from updates 25, 100, 200, and 250, then draw a task-stratified sample of 800 matched turns: 100 from each single-object family and 300 from Pick Two. For the structured-state matcher, we deduplicate exact matches and obtain a separate stratified sample of 781 turns. The two samples are therefore matcher-specific rather than paired row by row. In both audits, we replay the realized student action prefix to restore the reached TextWorld state, then execute the selected candidate followed by the remaining canonical suffix.

History replay succeeds on 792 of 800 turns. Its eight failures are confined to Look (2) and Pick Two (6). The Look failures arise when similar lamp-use observations conceal whether the target object is in inventory; the Pick Two failures arise when a visually similar cabinet state follows different object-level progress. In every case the selected candidate itself is admissible, but the subsequent canonical suffix does not complete the task. This is precisely the distinction between matching the current observation and matching execution progress.

All 781 structured-state replays succeed. On the audited matched turns, the structured matcher used by the reported ALFWorld runs thus reconnects the reached state to a complete successful continuation.

\subsection{WebShop Coverage}
\paragraph{WebShop rollout audit.}
We apply the same goal-progress matcher offline to 31,296 archived training turns from 30 checkpoints (updates 5--150) of the reported Qwen2.5-3B WebShop \method{} run. Table~\ref{tab:supp_webshop_audit} reports coverage by the parsed page type. Every routed candidate or candidate set is drawn from the current parsed admissible-action list. As a complementary set-guidance check, among 1,604 successful option-set turns, the sampled action belongs to the rendered compatible set on 1,525 (95.1\%). Relative to canonical-exact state matching on the same rows, goal-progress matching raises coverage from 49.4\% to 53.9\% and loses no canonical-exact match.

\begin{table}[t]
\centering
\setlength{\tabcolsep}{3.1mm}
\begin{tabular}{lrr}
\toprule
Page type & Turns & Matched \\
\midrule
Search home & 4,676 & 4,676 (100.0\%) \\
Search results & 7,538 & 4,298 (57.0\%) \\
Item page & 18,019 & 7,881 (43.7\%) \\
Other turns & 1,063 & 0 (0.0\%) \\
\midrule
All & 31,296 & 16,855 (53.9\%) \\
\bottomrule
\end{tabular}
\caption{Offline WebShop goal-progress matcher audit over archived on-policy turns. A matched row has nonempty, currently admissible local guidance.}
\label{tab:supp_webshop_audit}
\end{table}

\subsection{FTRL Coverage}
For the FTRL exact matcher, the mean matched-turn ratio across all 207 training updates is $41.2\%$. Excluding the initial turn, whose empty information state can match the empty reference prefix, the mean is $27.3\%$. The final training batch has corresponding ratios of $36.5\%$ and $21.9\%$. The means weight each update equally. Turn-matching rates are distinct from the $2{,}181/2{,}215$ task-level reference availability reported in Appendix~\ref{sec:reference_construction}. They show that the observed-information representation supports reference guidance beyond the common initial state.

\section{Efficiency and Response Characteristics}

\subsection{Training Overhead}

\begin{table}[t]
\centering
\setlength{\tabcolsep}{1.4mm}
\begin{tabular}{lrrr}
\toprule
Component & Mean & Median & p95 \\
\midrule
Reference lookup & 0.017 & 0.016 & 0.022 \\
Signature & 0.066 & 0.059 & 0.142 \\
Match + ground. & 0.200 & 0.172 & 0.423 \\
Render + bookkeeping & 0.550 & 0.286 & 0.859 \\
\midrule
Total & 0.833 & 0.569 & 1.335 \\
\bottomrule
\end{tabular}
\caption{Single-thread ALFWorld CPU time per matched turn, in milliseconds. Reference-index construction is a one-time cost and is not included.}
\label{tab:supp_cpu_runtime}
\end{table}

\paragraph{Deployment cost.}
The reference index, adapter, matcher, summary, and candidate are confined to training, so deployment adds neither prompt tokens nor an inference component. During training, reference indexing, signature construction, matching, action grounding, and context rendering run on CPU before the synchronized teacher scores the sampled response.

\paragraph{Teacher-context tokens.}
Table~\ref{tab:supp_prompt_tokens} uses the Qwen3-1.7B tokenizer on the same 800 matched probe turns, separating privileged-context tokens from the complete teacher prompt. \method{} adds 20.96 context tokens on average over \fullpath{}, whereas Skill-SD uses a substantially longer context.

\begin{table}[!htbp]
\centering
\begingroup
\fontsize{10}{12}\selectfont
\setlength{\tabcolsep}{1mm}
\begin{tabular*}{\columnwidth}{@{\extracolsep{\fill}}lrrrr@{}}
\toprule
Method & Mean & Median & p95 & Max \\
\midrule
\multicolumn{5}{l}{\textit{Privileged context}} \\
\fullpath{} & 112.9 & 112 & 148 & 175 \\
\method{} & 133.8 & 134 & 161 & 186 \\
Skill-SD & 943.5 & 888 & 1,174 & 1,174 \\
\midrule
\multicolumn{5}{l}{\textit{Full teacher prompt}} \\
\fullpath{} & 635.7 & 587 & 1,032 & 1,351 \\
\method{} & 656.6 & 616 & 1,041 & 1,390 \\
Skill-SD & 1,466.3 & 1,454 & 1,828 & 2,154 \\
\bottomrule
\end{tabular*}
\endgroup
\caption{Teacher-input tokens on 800 matched ALFWorld turns. Full prompts include the ordinary prompt and privileged context.}
\label{tab:supp_prompt_tokens}
\end{table}

\paragraph{CPU runtime.}
Table~\ref{tab:supp_cpu_runtime} profiles 2,400 calls on an Intel Xeon Platinum 8468V using one thread: three repeats of 800 matched turns after warm-up. Reference-index construction takes 437.1\,ms once and is excluded from per-turn times. Context reconstruction agrees with the frozen probe inputs.

\subsection{Response Length and Style}
\label{sec:response_style}

\paragraph{Response audit.}
We parse all turns from the fixed Qwen3-1.7B ALFWorld validation set (128 tasks and four rollouts per task) at the last checkpoint. Table~\ref{tab:supp_response_style_final} summarizes response length, repeated 4-grams, repeated normalized segments, and invalid actions. \method{} remains close to GRPO in response length, while \fullpath{} and especially Skill-SD are longer and more repetitive.

\begin{table}[!htbp]
\centering
\setlength{\tabcolsep}{0.9mm}
\begingroup
\fontsize{10}{12}\selectfont
\begin{tabular}{@{}lrrrr@{}}
\toprule
Method & Tokens & Rep.-4g & Rep.-seg. & Invalid \\
\midrule
\method{} & 78.8 & 8.8\% & 0.1\% & 0.1\% \\
\fullpath{} & 142.6 & 20.7\% & 1.0\% & 0.0\% \\
GRPO & 79.5 & 3.8\% & 0.0\% & 0.0\% \\
Skill-SD & 255.6 & 38.6\% & 9.2\% & 0.0\% \\
\bottomrule
\end{tabular}
\endgroup
\caption{Response statistics over all validation turns at the last checkpoint. Repeated 4-gram and segment rates are turn-level means; invalid actions use the environment validity field.}
\label{tab:supp_response_style_final}
\end{table}

\end{document}